\documentclass{article} 
\usepackage[preprint]{colm2025_conference}

\usepackage{microtype}
\usepackage{hyperref}
\usepackage{url}
\usepackage{booktabs}

\usepackage{lineno}
\usepackage{arydshln}
\usepackage{commenting}
\usepackage{tcolorbox}
\usepackage{multirow}
\usepackage{amsmath}
\usepackage{booktabs}

\definecolor{darkblue}{rgb}{0, 0, 0.5}
\hypersetup{colorlinks=true, citecolor=darkblue, linkcolor=darkblue, urlcolor=darkblue}

\usepackage{algorithm}
\usepackage{algpseudocode}
\usepackage{graphicx}
\usepackage{makecell}
\usepackage{fontawesome}
\usepackage{inconsolata} 
\usepackage{enumitem}
\usepackage{graphicx}
\usepackage{subcaption}
\usepackage{wrapfig}
\usepackage{afterpage}
\usepackage{placeins}
\usepackage{textcomp}
\usepackage{longtable}
\usepackage{selectp}
\usepackage{floatflt}
\usepackage{listings}
\usepackage{xcolor}
\usepackage{adjustbox}
\usepackage{pgffor}
\title{Understanding Aha Moments: from External Observations to Internal Mechanisms}

\author{
Shu Yang$^{1,2}$ \
Junchao Wu$^{3}$ \
Xin Chen$^{3}$ \
Yunze Xiao$^{4}$ \
\textbf{Xinyi Yang}$^{3}$ \\
\textbf{Derek F. Wong}$^{3}$ \
and \textbf{Di Wang}\thanks{Corresponding author.}$^{*,1,2}$ \\
$^1$Provable Responsible AI and Data Analytics (PRADA) Lab\\
$^2$King Abdullah University of Science and Technology\\
$^3$University of Macau \quad $^4$Carnegie Mellon University
}

\begin{document}

\ifcolmsubmission
\linenumbers
\fi

\maketitle

\begin{abstract}


Large Reasoning Models (LRMs), capable of reasoning through complex problems, have become crucial for tasks like programming, mathematics, and commonsense reasoning. However, a key challenge lies in understanding how these models acquire reasoning capabilities and exhibit ``aha moments'' when they reorganize their methods to allocate more thinking time to problems. In this work, we systematically study ``aha moments'' in LRMs, from linguistic patterns, description of uncertainty, ``Reasoning Collapse'' to analysis in latent space. 
We demonstrate that the ``aha moment'' is externally manifested in a more frequent use of anthropomorphic tones for self-reflection and an adaptive adjustment of uncertainty based on problem difficulty. This process helps the model complete reasoning without succumbing to ``Reasoning Collapse''. Internally, it corresponds to a separation between anthropomorphic characteristics and pure reasoning, with an increased anthropomorphic tone for more difficult problems. Furthermore, we find that the ``aha moment'' helps models solve complex problems by altering their perception of problem difficulty. As the layer of the model increases, simpler problems tend to be perceived as more complex, while more difficult problems appear simpler. 
\end{abstract}

\section{Introduction}

Large reasoning models (LRMs), which can ``think through'' complex problems, are emerging as a new class of large-scale models that enhance reliability and performance in tasks such as programming, mathematics, and commonsense reasoning ~\citep{guo2025deepseek,xu2025towards}. 
Recent studies have shown that reasoning capabilities can be incentivized through reinforcement learning, leading to an ``aha moment'' where the model suddenly exhibits the ability to rethink a problem using an anthropomorphic tone. Additionally, these reasoning abilities can be distilled from a well-trained LRM into a smaller LLM~\citep{guo2025deepseek,meng2025mmeurekaexploringvisualaha}.
While we are all inspired by this, our understanding remains limited regarding how the ``aha moment''— where the model allocates more thinking steps to a problem by reflecting on its initial solution with an anthropomorphic tone (as shown in Figure~\ref{fig:ahaexample})—can help push the limits of reasoning ability. 

To fill the gap, our work analyzes the mechanism of the ``aha moment,'' moving from surface-level linguistic patterns and uncertainty to deeper latent space exploration. We find that the ``aha moment'' influences the model's external behavior in three key ways: (1) it leads to the emergence of more anthropomorphic linguistic patterns, (2) it enables the model to adjust its uncertainty based on the difficulty level of a given problem, promoting more self-reflection organized through anthropomorphic language, (3) it can prevent the model from falling into Reasoning Collapse, where the model fails to complete reasoning correctly due to language mixing, token sequence repetition, or repeated reasoning paths.  

Building on these observations, we propose the Reasoning-Anthropomorphic Separation Metric (RASM), a quantitative measure that captures the shift between anthropomorphic self-reflection and pure reasoning in model outputs. Our findings reveal that the ``aha moment'' is intrinsically linked to a model’s ability to dynamically adjust its thinking strategy based on task difficulty, with anthropomorphic traits becoming more pronounced as problems grow more complex. This suggests that incorporating human-like expressions can help enhance a model’s capacity with organized self-reflection and navigate challenging reasoning tasks. To gain deeper insight into how the “aha moment” aids LLMs in solving more complex questions, we analyze the internal mechanisms underlying these reasoning behavior shifts through latent space exploration. 
Our analysis of internal activations uncovers an interesting trend in R1-like models: while they initially distinguish problem difficulty levels clearly, deeper layers progressively \textit{blur these boundaries}. This phenomenon contrasts sharply with standard reasoning processes, where a model’s understanding of problem difficulty typically becomes clearer as layer depth increases. This indicates that ``aha moment'' can help the R1-like model's reasoning with processing simple problems in a more complex manner while interpreting complex problems as simpler ones. This finding further supports the ``overthinking'' phenomenon~\citep{chen2024not, sui2025stopoverthinkingsurveyefficient} in LRMs, where models tend to produce overly verbose and redundant responses even for simple questions.
Overall, our results highlight the critical importance of investigating both reasoning strategies and anthropomorphic tendencies in R1-like models.

Our work inspires further exploration into optimizing R1-like models—not only by improving reasoning strategies but also by enhancing their anthropomorphic behaviors and refining their ability to recognize task difficulty.



\begin{figure}[t]
    \centering
    \vspace{-0.2cm}
    \resizebox{\textwidth}{!}{
        \scriptsize
        \begin{tabular}{p{0.47\textwidth}  p{0.63\textwidth}}
            \toprule
            \multicolumn{2}{l}{\small \textbf{Question} What is the coefficient of $x^2y^6$ in the expansion of $\left(\frac{3}{5}x-\frac{y}{2}\right)^8$? Express your answer as a common fraction.}\\
            \midrule
            \small \textbf{Original (no-aha) Model Response} & \small \textbf{Reasoning (aha) Model Response} \\
            \midrule
            \textbf{Qwen2.5-Math-1.5B:} 
            
            \textcolor{myblue}{To find the} coefficient of ..., \textcolor{myblue}{we can} use the binomial theorem. The binomial theorem states that: \( (a + b)^n = \sum_{k=0}^{n} \binom{n}{k} a^{n-k} b^k\). 
            \textcolor{myblue}{In this case},..., \textcolor{myblue}{so we need to} find the term in the expansion...
            \textcolor{myblue}{Simplifying this} term, we get: ...
            
            \textcolor{myblue}{So} the coefficient of \(x^2 y^6\) in the expansion of \(\left(\frac{3}{5}x - \frac{y}{2}\right)^8\) is \(\frac{63}{400}\).
            
            \textcolor{myblue}{Let's confirm this} with Python code...
            
            \texttt{<python code and output>}
            
            The coefficient of \(x^2 y^6\) in the expansion of \(\left(\frac{3}{5}x - \frac{y}{2}\right)^8\) is \(\frac{63}{400}\), which is \textcolor{myblue}{equivalent to} 0.1575. \textcolor{myblue}{So the final answer is}: \( \frac{63}{400} \)
            &
            \textbf{DeepSeek-R1-Distill-Qwen-1.5B:}
            
            \colorbox{myorange!40}{\texttt{<think>}} 
            \textcolor{myorange}{Okay, so I need} to find the coefficient of \( x^2 y^6 \)... \textcolor{myorange}{Hmm, alright, let me think about} how to approach this.  \textcolor{myorange}{I remember something} about.... \textcolor{myorange}{Yeah}, the binomial theorem states that \( (a + b)^n = \sum_{k=0}^{n} \binom{n}{k} a^{n - k} b^k \). \textcolor{myblue}{So,} ...
            \textcolor{myorange}{But wait}, ... \textcolor{myblue}{But that's okay because} it's just ... , the term ...
            \textcolor{myorange}{Okay, so I need} to compute ... \textcolor{myblue}{Now, simplify} \( \frac{252}{1600} \). \textcolor{myorange}{Let me see if I can} reduce this fraction.
            \textcolor{myorange}{Wait, just to make sure I didn't make any mistakes in my} ...
            \textcolor{myblue}{**Final Answer**}
            The coefficient of \( x^2 y^6 \) is \( \frac{63}{400} \) .
            
            \colorbox{myorange!40}{\texttt{</think>}}
            
            \textcolor{myblue}{To find the} coefficient of ...,  we use the binomial theorem. The binomial theorem states that: ... \textcolor{myblue}{First, compute} the binomial coefficient ...
            \textcolor{myblue}{Next, compute} ...
            \textcolor{myblue}{Then, compute} ...
            \textcolor{myblue}{Now, multiply all}Now, multiply all parts together ...
            \textcolor{myblue}{Simplify the constants} ...
            \textcolor{myblue}{Thus,} the coefficient of \( x^2 y^6 \) is \( \frac{63}{400} \).
            \\
            \midrule
            \textbf{Llama-3.1-8B-Instruct:}
            
            \textcolor{myblue}{To find the} coefficient of $x^2y^6$ in the expansion of $\left(\frac{3}{5}x-\frac{y}{2}\right)^8$, we can use the Binomial Theorem. The Binomial Theorem states that for any non-negative integer $n$ and any real numbers $a$ and $b$, we have $(a+b)^n = \sum_{k=0}^{n} \binom{n}{k}a^{n-k}b^k.$
            \textcolor{myblue}{In this case, we have} $a = \frac{3}{5}x$, $b = -\frac{y}{2}$, and $n = 8$. We want to find the coefficient of $x^2y^6$, \textcolor{myblue}{so we need to} find the term in the expansion where the power of $x$ is 2 and the power of $y$ is 6.
            \textcolor{myblue}{Using the Binomial Theorem}, we have ...
            \textcolor{myblue}{We want to} find the term where $k = 6$, \textcolor{myblue}{so we have} ...
            \textcolor{myblue}{so we have ...} \textcolor{myblue}{Evaluating this expression, we get} ...
            \textcolor{myblue}{Simplifying, we get} ...
            \textcolor{myblue}{Therefore}, the coefficientof $x^2y^6$ in the expansion of $\left(\frac{3}{5}x-\frac{y}{2}\right)^8$ is $\frac{252}{1600} ==\frac{63}{400}$.
            &
            \textbf{DeepSeek-R1-Distill-Llama-8B:}
            
            \colorbox{myorange!40}{\texttt{<think>}}
            \textcolor{myorange}{Okay, so I need} to find the coefficient of ... \textcolor{myorange}{Hmm, let me think about} how to approach this.
            \textcolor{myorange}{I remember that} ... \textcolor{myblue}{binomial theorem} ...
            \textcolor{myblue}{In this case}, the binomial is ..., \textcolor{myblue}{so} ...
            \textcolor{myorange}{I need the term} where the power of ...
             \textcolor{myorange}{Let me denote} \textcolor{myblue}{the general term as} ...
             \textcolor{myorange}{Let me} \textcolor{myblue}{compute that term}. \textcolor{myblue}{First, calculate} ... 
            \textcolor{myblue}{Next, calculate} ... 
            \textcolor{myblue}{Then, calculate} ...  
            \textcolor{myblue}{Now, multiply all} these together: ...
            \textcolor{myorange}{Wait, let me double-check my steps} ...
            \textcolor{myorange}{So, I think} that's correct.
            \textcolor{myblue}{**Final Answer**}
            The coefficient of \( x^2 y^6 \) is \(\frac{63}{400}\).
            
            \colorbox{myorange!40}{\texttt{</think>}}
            
            To find the coefficient of ...  \textcolor{myblue}{we use the binomial theorem} ...
            \textcolor{myblue}{1. Calculate the} binomial coefficient ...
            \textcolor{myblue}{2. Calculate} ...
            \textcolor{myblue}{3. Calculate} ...
            \textcolor{myblue}{4. Multiply these} ...
            \textcolor{myblue}{5. Simplify} the fraction ...
            \textcolor{myblue}{Thus}, the coefficient of \( x^2 y^6 \) is \(\frac{63}{400}\).
            \\  
            \bottomrule
        \end{tabular}
    }
    \caption{\textbf{Example of an "Aha Moment."} We highlight the \textcolor{myblue}{reasoning steps} and \textcolor{myorange}{anthropomorphic expressions} separately. This example illustrates how the ``aha moment'' integrates structured reasoning with the emergence of anthropomorphic language patterns. The question is sourced from the GSM8K dataset.}
    \label{fig:ahaexample}
    \vspace{-0.3cm}
\end{figure}





\vspace{-0.1in}
\section{Preliminaries}
\vspace{-0.1in}
\subsection{Synthetic Puzzle Dataset}
\label{data:model and synthetic data}
\vspace{-0.1in} 
\begin{wrapfigure}{r}{0.45\textwidth}
    \centering
     \vspace{-0.2cm}
    \small
    \resizebox{0.47\textwidth}{!}{
        \begin{tabular}{@{}p{4cm} p{4.5cm}}
        \toprule
        \textcolor{myblue}{\textbf{No-aha models}} & \textcolor{myorange}{\textbf{Corresponding aha models}} \\ \midrule
        Qwen2.5-Math-1.5B-Instruct & DeepSeek-R1-Distill-Qwen-1.5B \\
        Qwen2.5-Math-7B-Instruct & DeepSeek-R1-Distill-Qwen-7B \\
        Llama-3.1-8B-Instruct & DeepSeek-R1-Distill-Llama-8B \\ 
        Qwen2.5-14B-Instruct & DeepSeek-R1-Distill-Qwen-14B \\
        \bottomrule
        \end{tabular}
    }
    \caption{No-aha models and their corresponding aha models we used.}
    \label{tab:models}
    \vspace{-0.2cm}
\end{wrapfigure}
To avoid the influence of varying difficulties, task structures, and potential data leakage that could affect the fairness of our experiments and results, we generate synthetic reasoning tasks motivated by previous works~\citep{xie2024memorization,tinyzero}. In the following tasks, we use $n$ to define the difficulty level of the puzzles. Specifically, for common sense reasoning, we employ Knights and Knaves (K\&K) puzzles; for mathematical reasoning, we use the Count Down puzzle. Next, we provide examples and a detailed description of these two tasks.
\paragraph{Knights and Knaves (K\&K)}\citep{JOHNSONLAIRD199069} is a classic logical puzzle that requires determining which individuals are telling the truth and which are lying based on the statements of $n$ people,  and then identifying their true identities.

\begin{tcolorbox}[
    colframe=myorange!90!black, 
    colback=myorange!5!white,
    coltitle=white,
    fonttitle=\bfseries, 
    title= \scriptsize  An example of a K\&K puzzle  (\ensuremath{n = 3}),
    boxrule=0.5mm, 
]
\scriptsize 
\textbf{Problem}:  
A very special island is inhabited only by knights and knaves. Knights always tell the truth, and knaves always lie. You meet 3 inhabitants: Penelope, David, and Zoey. Penelope noted, ``David is a knight if and only if David is a knave''. David told you that ``Zoey is a knave if and only if Zoey is a knight''. According to Zoey, ``If Penelope is a knave then David is a knave''. So who is a knight and who is a knave? \\
\textbf{Solution}:  
Penelope is a knave, David is a knave, and Zoey is a knight.
\label{dataexample: kkpuzzle}
\end{tcolorbox}

\textbf{Count Down} \citep{tinyzero} is a preprocessed dataset for the countdown task, where the goal is to generate equations to reach a given target number using \textit{n} provided numbers.
\begin{tcolorbox}[
    colframe=myorange!90!black, 
    colback=myorange!5!white,
    coltitle=white,
    fonttitle=\bfseries, 
    title= \scriptsize  An example of a Count Down puzzle  (\ensuremath{n =3}),
    boxrule=0.5mm, 
]
\scriptsize
\textbf{Problem}:  
Using the numbers \texttt{\{2, 26, 38\}}, create an equation that equals \texttt{90}. You can use basic arithmetic operations (\texttt{+, -, *, /}) and each number can only be used once.

\textbf{Solution}: \texttt{2*26+38}
\label{puzzle}
\end{tcolorbox}

Compared to K\&K, where the model heavily relies on understanding and distinguishing the given information, the Count Down task provides much more limited information. As a result, the model must continuously explore and engage in mathematical reasoning to arrive at the correct answer.

\subsection{The ``Aha Moment'' Phenomenon}

A recent breakthrough in LRMs, DeepSeek R1~\citep{guo2025deepseek}, demonstrated that reinforcement learning (RL) can enhance a model’s reasoning abilities without requiring any supervised reasoning data. The most intriguing aspect of their findings is the discovery of an ``aha moment''—when applying RL with a rule-based reward, the model autonomously developed advanced problem-solving strategies by reflection and self-correction in an \textit{anthropomorphic tone} (as shown in Figure \ref{fig:ahaexample}), leading to improved performance on reasoning tasks. 

Inspired by this phenomenon, numerous studies have attempted to reproduce the R1-Zero-like training paradigm. Most of these works observed that after RL training, LLMs began exhibiting self-reflection behaviors when answering a question. However, \citet{liu2025oatzero} found instances where self-reflection occurred as early as epoch 0, suggesting that ``achieving an aha moment'' cannot simply be defined based on reflection behavior. Similarly, in \citet{xie2025logicrlunleashingllmreasoning}'s replication of R1-like training, the model demonstrated complex self-correct reasoning behaviors during training but did not explicitly verbalize an anthropomorphic tone (e.g., ``Wait, wait. Wait. That's an aha moment...'') as observed in~\cite{guo2025deepseek}. Based on this exploration, to better understand the mechanisms underlying the aha moment, we need not only to learn about different reasoning behaviors but also to explore the role of anthropomorphic tone in reasoning tasks. Therefore, we analyze both model behavior and internal latent space representations using two syntactic puzzle datasets, K\&K and Count Down, as detailed in \S~\ref{data:model and synthetic data}.

\noindent \textbf{``Aha'' and ``no-aha'' models}~In this paper, we use the DeepSeek-R1-Distill models ranging from 1.5B to 14B as the ``aha models,'' which are distilled from the DeepSeek-R1 model and retain certain R1-like characteristics. Correspondingly, we use their respective Instruct models released by the Qwen or Llama team~\citep{qwen2.5,grattafiori2024llama3herdmodels} as the ``no-aha models,'' rather than the base models before distillation fine-tuning. This choice prevents potential unfairness in evaluation, as using the base models before instruction fine-tuning could result in an inability to properly follow our provided questions. Table~\ref{tab:models} lists the aha models alongside their corresponding no-aha models.
\section{Analyzing the ``Aha Moment'' through Linguistic Pattern and Uncertainty}
\label{sec:lingua and uncertainty}
To investigate how the ``aha moment'' helps models in completing a reasoning task, we first analyze the differences in linguistic patterns between aha and no-aha model pairs in Table~\ref{tab:models} (\S~\ref{subsec:linguistic pattern}). We then explore the profound significant impact of these linguistic patterns on LLMs from an uncertainty perspective, which is highly related to self-reflection or self-correction behaviors (\S~\ref{subsec: uncertainty}). Finally, we further discover that ``aha moment'' can prevent models from falling into ``reasoning collapse,'' where they struggle to reach a final answer due to ``language mixing,'' ``language repetition,'' and ``reasoning path repetition'' (\S~\ref{subsec:Reasoning collapse}). This suggests that more anthropomorphism and uncertainty may help models escape from reasoning failure.  
    

     

\begin{figure*}[h]
    \centering
    \includegraphics[width=0.40\textwidth, trim=0.2cm 0.2cm 0.2cm 0.2cm, clip]{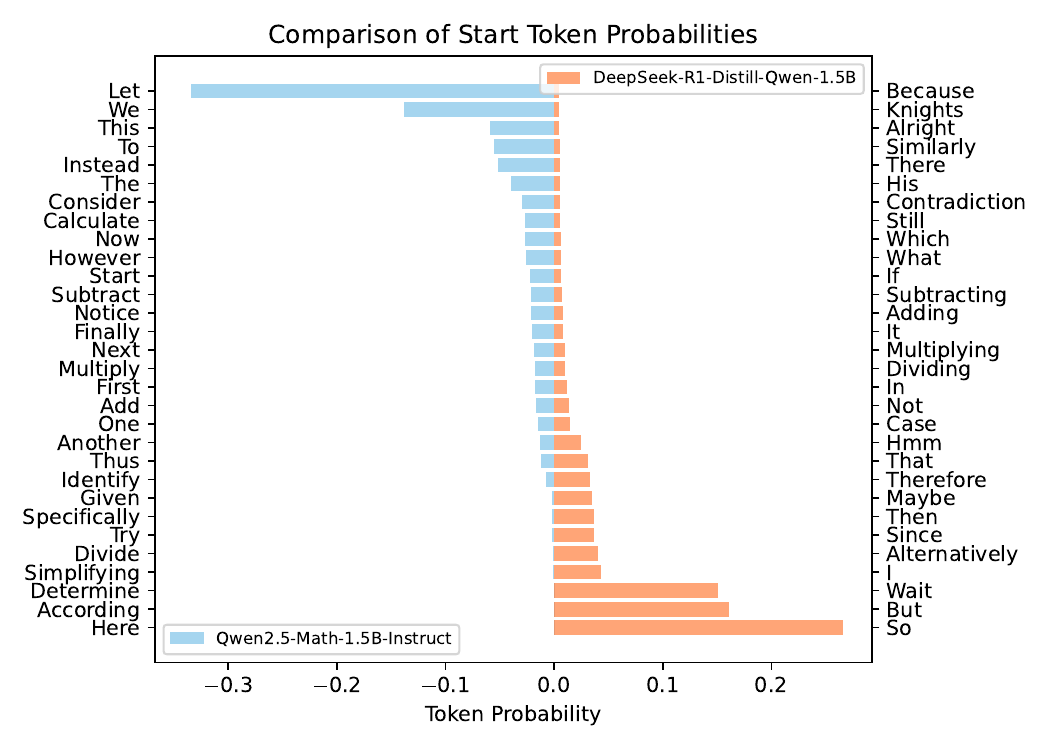}
    \hspace{0.15cm}
    \includegraphics[width=0.40\textwidth, trim=0.2cm 0.2cm 0.2cm 0.2cm, clip]{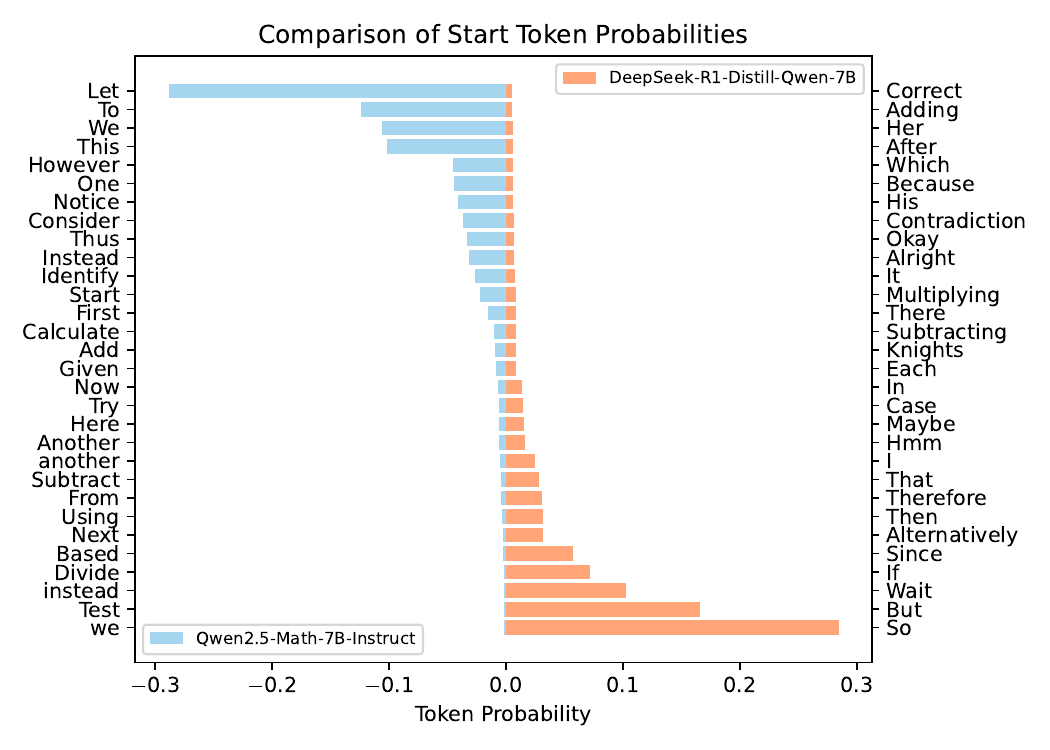}
    \hspace{0.15cm}
    \includegraphics[width=0.40\textwidth, trim=0.2cm 0.2cm 0.2cm 0.2cm, clip]{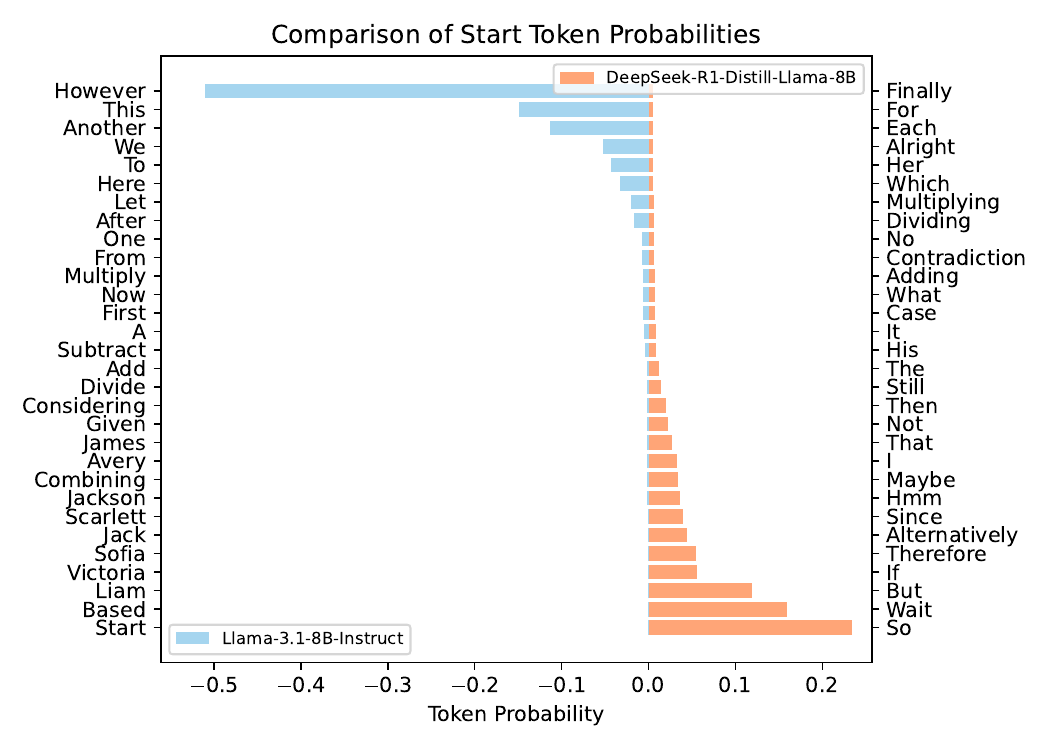}
    \hspace{0.15cm}
    \includegraphics[width=0.40\textwidth, trim=0.2cm 0.2cm 0.2cm 0.2cm, clip]{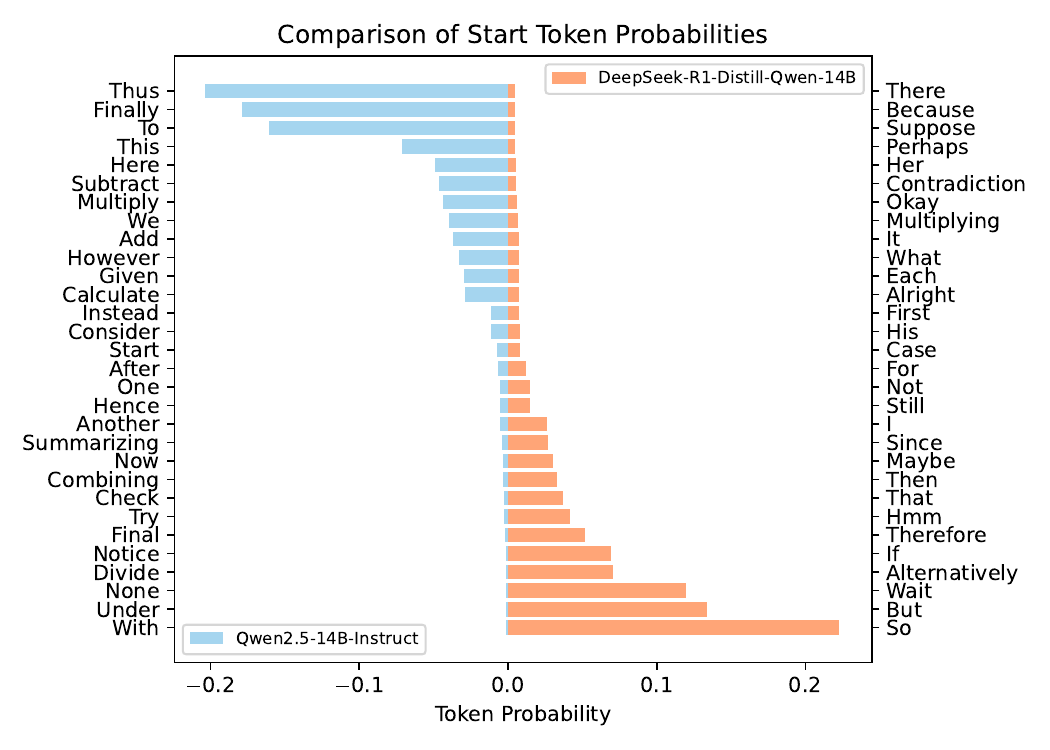}
\caption{The beginning token distribution of \textcolor{myblue}{no-aha model} and \textcolor{myorange}{aha model} pairs}
\vspace{-0.3cm}
\label{fig:token_frequency}
\end{figure*}

\subsection{Different Beginning Token Distribution of Aha Model and No-aha Model} 
\label{subsec:linguistic pattern}

Previous research has highlighted a strong correlation between the beginning token of generated text and model performance. By providing appropriate prefix tokens, the model's decision space can be effectively constrained, enhancing specific capabilities~\citep{DBLP:conf/acl/ZhanYWC024, ma2024focused}. In the context of reasoning tasks, \citet{ji2025first} found that the beginning tokens of different logical steps in multi-solution generation exhibit high consistency and can be leveraged for efficient fine-tuning. This underscores the critical role these beginning tokens play as an important linguistic pattern in shaping the model's reasoning performance. Inspired by this, we analyze the beginning tokens in the output of the aha models and their corresponding no-aha models. As shown by the example in Figure \ref{fig:ahaexample}, the output of the no-aha model typically unfolds with \textit{fixed reasoning structures} like ``To find the...'', while the aha model more often employs \textit{anthropomorphic tone expressions} such as ``Okay, so I need...'', forming two well-distinct linguistic patterns. To systematically analyze these differences, we sample 100 questions from each difficulty level of the K\&K and Count Down puzzle datasets and compute the distribution of beginning tokens in each sentence of both model types' outputs. To ensure a more meaningful comparison, we exclude high-frequency stop words common to both models such as \texttt{**}'' and \texttt{**\string\n}''.

The results, shown in Figure~\ref{fig:token_frequency}, reveal significant differences in linguistic patterns between aha and no-aha models. Aha models tend to adopt conversational and emotional expressions (e.g., ``Wait,'' ``Alright,'' ``Hmm,'' ``Okay,''), making their linguistic style more similar to natural human expression. They exhibit strong internal thinking patterns and display anthropomorphic characteristics. Especially, aha models prefer to use ``I'' instead of ``we'' compared to no-aha models, suggesting greater self-awareness. In contrast, no-aha models emphasize logical and structured expressions (e.g., ``Combining,'' ``According,'' ``Determine,'' ``First'') that directly relate to the problem-solving process, resulting in a more mechanical response. We also observed that, despite being of nearly the same scale, the Qwen2.5-Math-7B-Instruct and LLaMA-3.1-8B-Instruct models exhibit different preferences in linguistic patterns. Specifically, the LLaMA-3.1-8B-Instruct model tends to use personal names such as ``Divide'' and ``James'' more frequently as sentence-beginning tokens (or as the subjects of the sentence) compared to the Qwen model. This phenomenon has also been noted by~\citet{cheng2024anthroscore}, suggesting that LLaMA-3.1-8B-Instruct demonstrates a potential \textit{tendency} toward anthropomorphism. However, all no-aha models consistently fail to demonstrate the obvious self-awareness characteristic as their aha counterparts, allowing us to identify this anthropomorphic tone (or self-awareness) as a distinguishing feature between paired ``aha'' and ``no-aha'' models.


\subsection{Aha Models Dynamically Adjust their Uncertainty to Difficulty Levels}
\label{subsec: uncertainty}


\begin{figure}[h]
 \vspace{-0.3cm}
    \begin{subfigure}[b]{0.61\linewidth}
        \includegraphics[width=\linewidth, trim=0 0 0 0, clip]{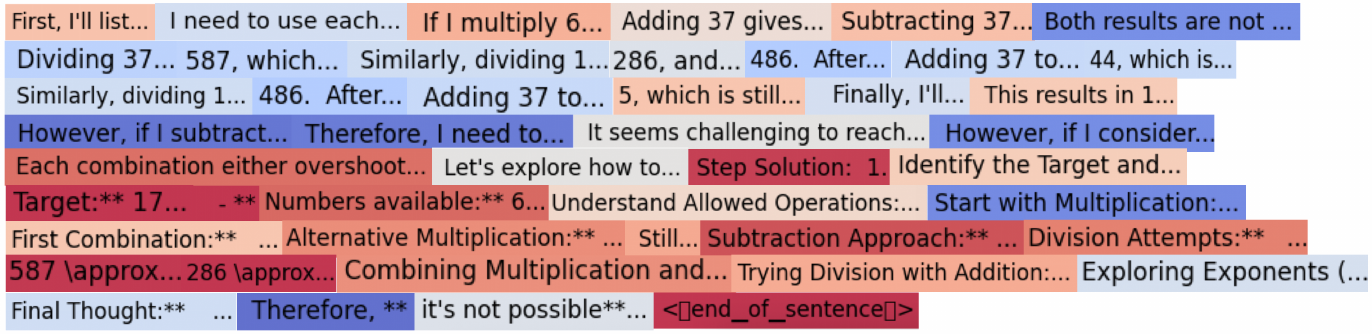}
        \subcaption{} 
        \label{subfig:uncertainty_a}
    \end{subfigure}
    \begin{subfigure}[b]{0.39\linewidth}
        \includegraphics[width=\linewidth, trim=0 0 0 0, clip]{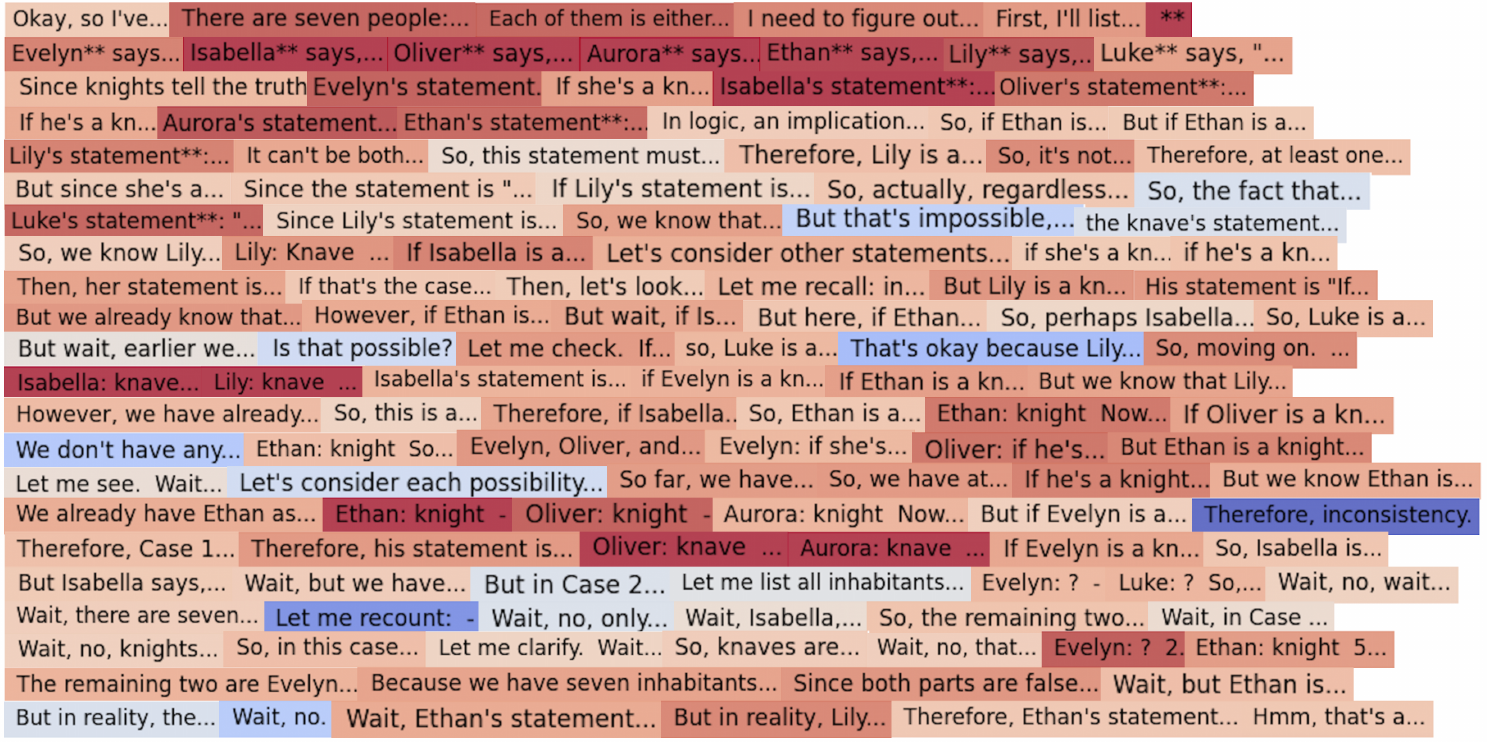}
        \subcaption{} 
        \label{subfig:uncertainty_b}
    \end{subfigure}
    \caption{Response examples from DeepSeek-R1-Distill-Qwen-1.5B in solving different tasks, Count Down (a)  and K\&K (b). Cooler colors indicate lower probability (higher uncertainty), while warmer colors indicate higher probability (lower uncertainty).}
    \label{fig:uncertainty_example}
     \vspace{-0.2cm}
\end{figure}

\begin{figure}
    \centering
    \vspace{-0.3cm}
    \includegraphics[width=0.80\linewidth, trim=0 0 0 0, clip]{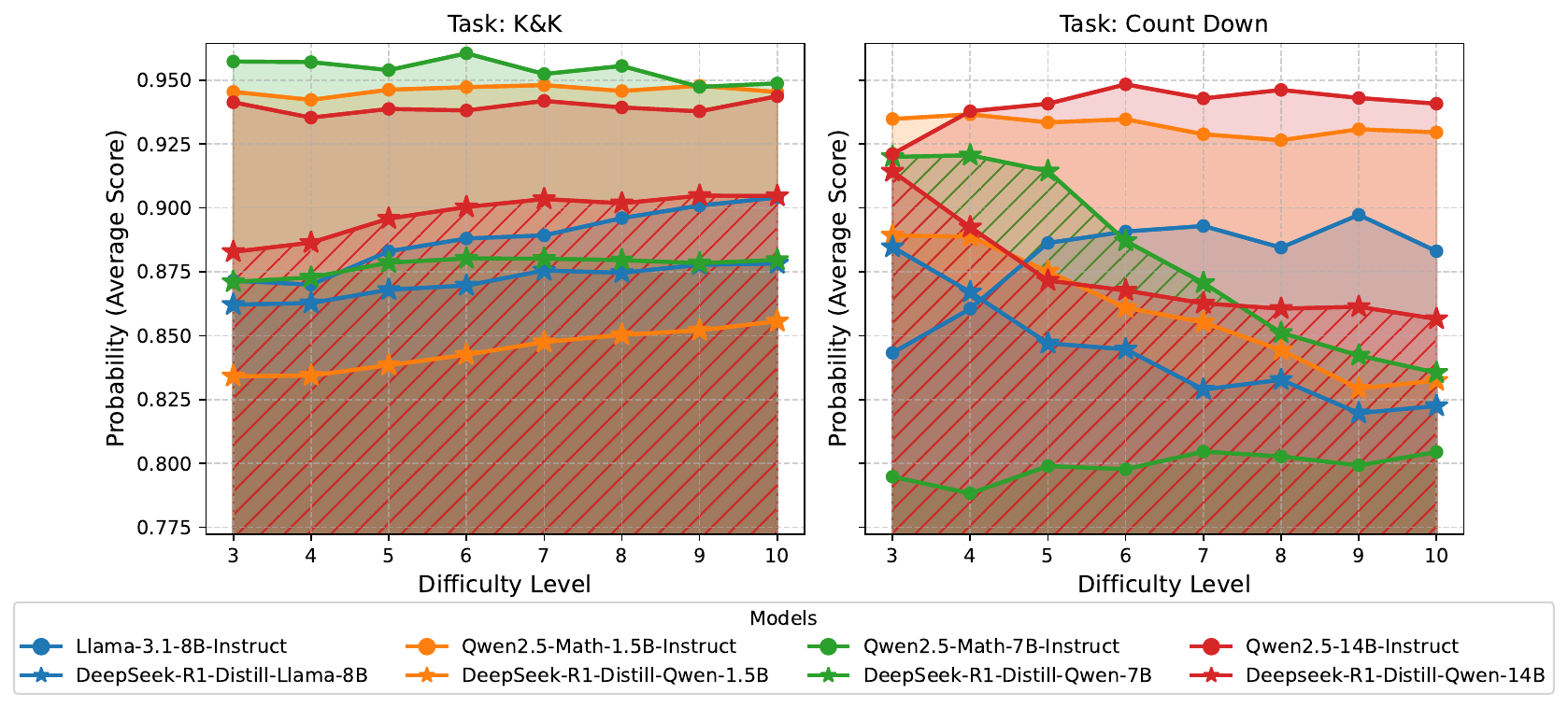}
    \caption{Average probability of LLMs' output across different task and difficulty level.}
    \label{fig:average_uncertainty}
    \vspace{-0.3cm}
\end{figure}


Having examined the distinct linguistic patterns between "aha" and "no-aha" models in §\ref{subsec:linguistic pattern}, this section investigates how these models express uncertainty in their inference process and how this uncertainty relates to increased self-reflection. We segment the LLMs' solution into individual sentences and compute the average probability of each sentence, as illustrated in Figure~\ref{fig:uncertainty_example}. We can find that when solving the Count down problem, the probability of output sentences gradually increases. Sentences containing an anthropomorphic tone or self-reflective expressions tend to have lower probabilities, indicating higher uncertainty. This pattern reflects the model’s reasoning process when tackling a mathematical problem with less information: it progressively refines its understanding through self-reflection, thereby increasing confidence in the final answer. However, when solving the K\&K problem, which involves numerous pieces of evidence in the input, the model frequently refers back to the provided information. Sentences that restate given details tend to have higher probabilities, indicating lower uncertainty. In contrast, sentences involving human-like self-reflection, such as ``Is that possible?'' or ``Wait,'' generally exhibit higher uncertainty. We provide a more detailed analysis of the most probable yet uncertain sentences for each model in Appendix~\ref{app:uncertain sentences}.

In Figure~\ref{fig:average_uncertainty}, we present the average probability of each model across different tasks and difficulty levels. Our analysis reveals that, as difficulty increases, the probability of sentences generated by the no-aha model does not exhibit a clear modified trend. Moreover, except for the performance of Qwen2.5-Math-7B-Instruct on the Count Down task, the no-aha model consistently maintains higher probability values than its corresponding aha model. This suggests that the no-aha model has lower uncertainty and may even struggle to identify the difficulty of a question, let alone dynamically adjust its reasoning path based on the difficulty level. In contrast, the uncertainty of the aha models follows a more consistent trend as difficulty varies. Specifically, in the K\&K task, the uncertainty of the aha model gradually decreases with increasing difficulty levels. This may be because the model repeatedly reiterates the conditions provided in the problem during reasoning, as illustrated in Figure~\ref{subfig:uncertainty_b}. For the Count Down task, the uncertainty of the aha model increases with difficulty, suggesting that aha models are capable of recognizing problems of varying difficulty or tasks and adjusting their responses accordingly. 

\subsection{The ``Reasoning Collapse'' of Aha and No-aha models}
\label{subsec:Reasoning collapse}
\begin{figure}[h]
\vspace{-0.2cm}
    \begin{subfigure}[b]{0.230\linewidth}
        \includegraphics[width=\linewidth, trim=0.7cm 0.2cm 0.7cm 0.7cm, clip]{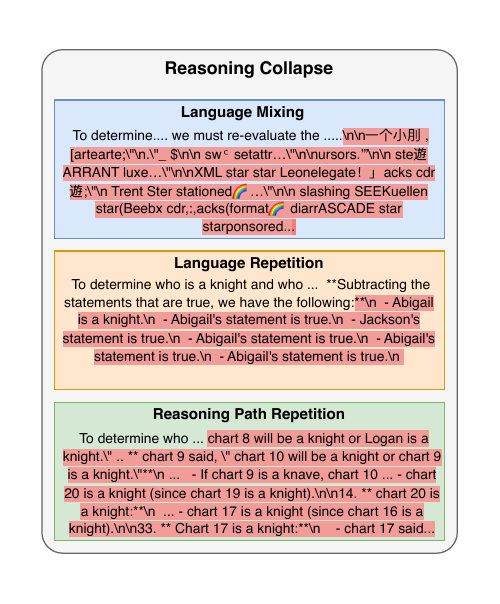}
        \subcaption{} 
        \label{fig:reasoning_collapse_example}
    \end{subfigure}
    \begin{subfigure}[b]{0.769\linewidth}
        \includegraphics[width=\linewidth, trim=0 0 0 0, clip]{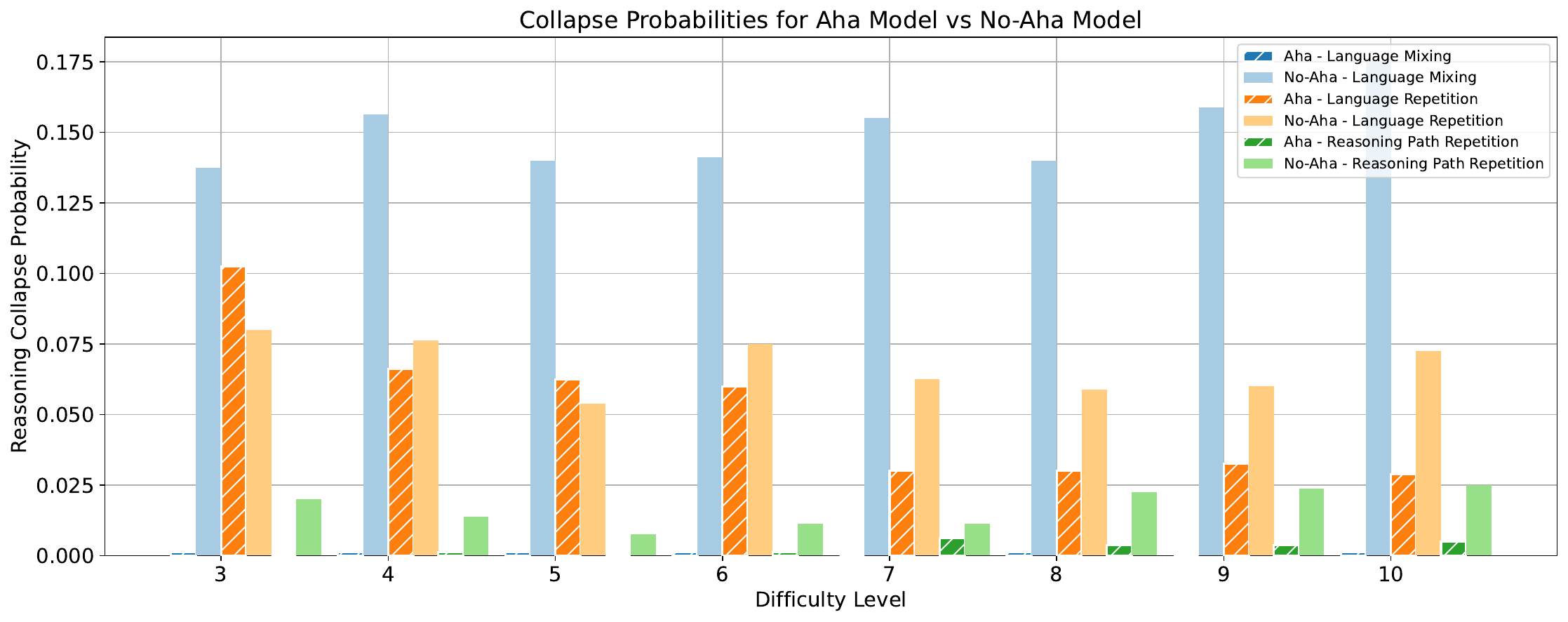}
        \subcaption{}
        \label{fig:reasoning_collapse}
    \end{subfigure}
    \caption{\textbf{Reasoning Collapse}: examples (a) and  occurrence probabilities across different model types and difficulty levels (b).}
    \label{fig:reasoning_collaspe_all}
    \vspace{-0.2cm}
\end{figure}

We also find that as problem difficulty increases, models exhibit more instances of \textbf{reasoning collapse}, which can be broadly categorized into three types: ``language mixing,'' ``language repetition,'' and ``reasoning path repetition.'' As illustrated in Figure~\ref{fig:reasoning_collapse_example}, ``language mixing'' can be defined as after generating several tokens, the model begins to produce a mixture of languages, sometimes including non-English characters or even emojis. ``Language repetition'' refers to cases where the model continuously repeats a sequence of tokens until it reaches the maximum output length. Finally, ``reasoning path repetition'' describes a situation in which the model becomes trapped in a fixed pattern of reasoning, unable to break out of the loop, leading to repetitive thought processes until the output reaches its maximum length.  We use non-English character detection to identify language mixing, consecutive repeated token and sequence detection for language repetition, and high-similarity phrase matching to capture reasoning path repetition. To quantify the likelihood of reasoning collapse, we calculate the occurrence probability of these patterns across 200 test samples at each difficulty level.

In Figure~\ref{fig:reasoning_collapse} we can demonstrate that aha models effectively suppress language mixing and reasoning path repetition, with both rates approaching zero, especially for the difficulty level 3-6. Additionally, when the problem difficulty level is low, the language repetition rate of aha models is larger than that of no-aha models. However, when the problem difficulty exceeds 5, aha models gradually gain an advantage. This may be because aha models are more prone to overthinking simple problems~\citep{chen2025think23overthinkingo1like,sui2025stopoverthinkingsurveyefficient}, leading to repeated descriptions of the same step, which results in higher language repetition. To further investigate the underlying reasons behind these phenomena in both the aha and no-aha models, we explore the ``aha moment'' through their latent space in the next section.

\section{Investigating the Aha Moment through Latent Speace Exploration}

In \S~\ref{sec:lingua and uncertainty}, we examined the differences in linguistic patterns, uncertainty and  ``reasoning collapse'' between aha and no-aha models during inference. In this section, we further investigate the ``aha moment'' by exploring the latent space of these models to understand how anthropomorphic features are activated (\S~\ref{subsec:rasm}) and how these models differ in reasoning the difficulty level of problems (\S~\ref{subsec:contrastiveanalysis}).
\subsection{Reasoning-Anthropomorphic Separation Metric}
\label{subsec:rasm}
\begin{figure}[htb]
    \centering
    \scriptsize
\begin{adjustbox}{width=1.0\textwidth}
\renewcommand{\arraystretch}{0.3}
    \begin{tabular}{@{\hspace{-0.15cm}}c@{\hspace{-0.15cm}}c@{\hspace{-0.15cm}}c@{\hspace{-0.15cm}}c@{\hspace{-0.15cm}}c@{\hspace{-0.15cm}}}
        $n = 3$ & $n = 4$ & $n = 5$ & $n = 7$ & $n = 10$ \\
        \hline
        \includegraphics[width=0.2\textwidth, trim=0.2cm 0.2cm 0.2cm 0.2cm, clip]{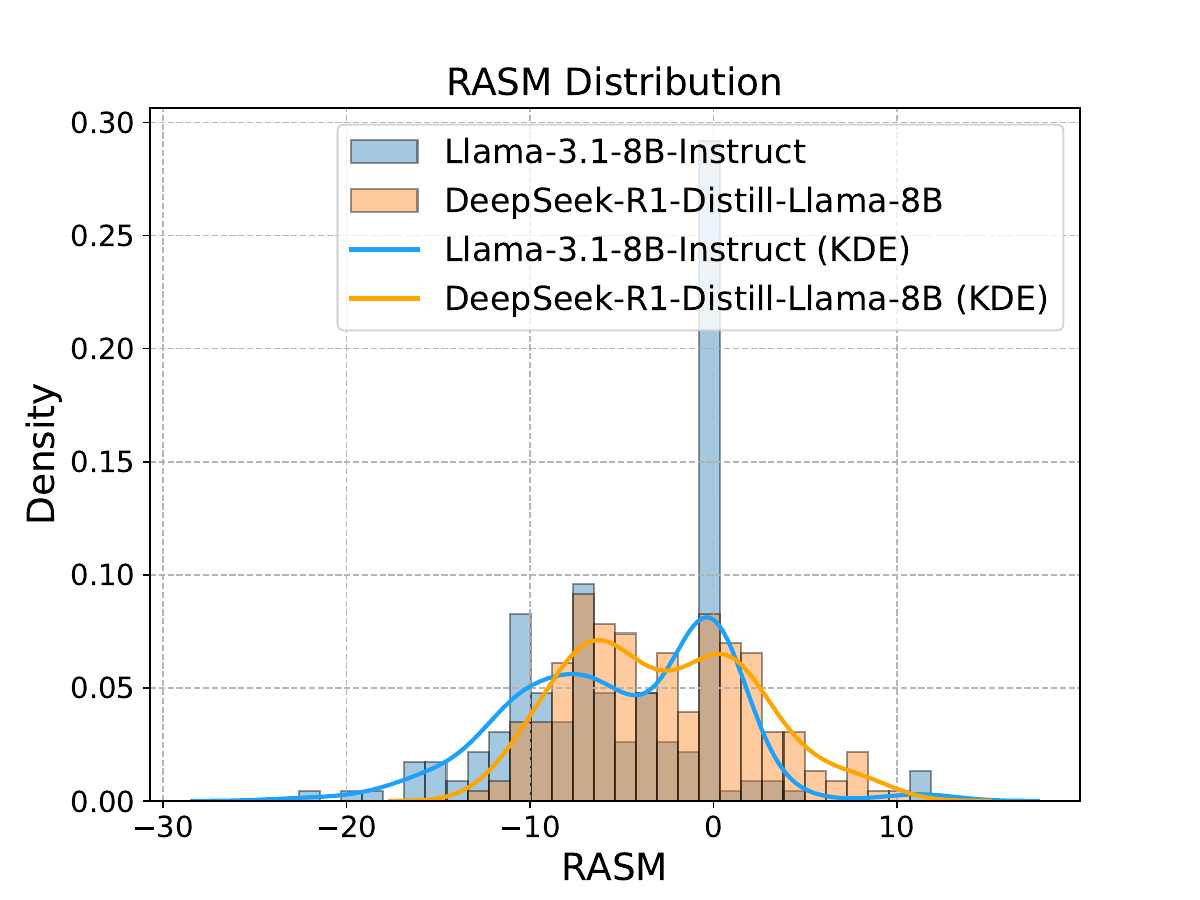} &
        \includegraphics[width=0.2\textwidth, trim=0.2cm 0.2cm 0.2cm 0.2cm, clip]{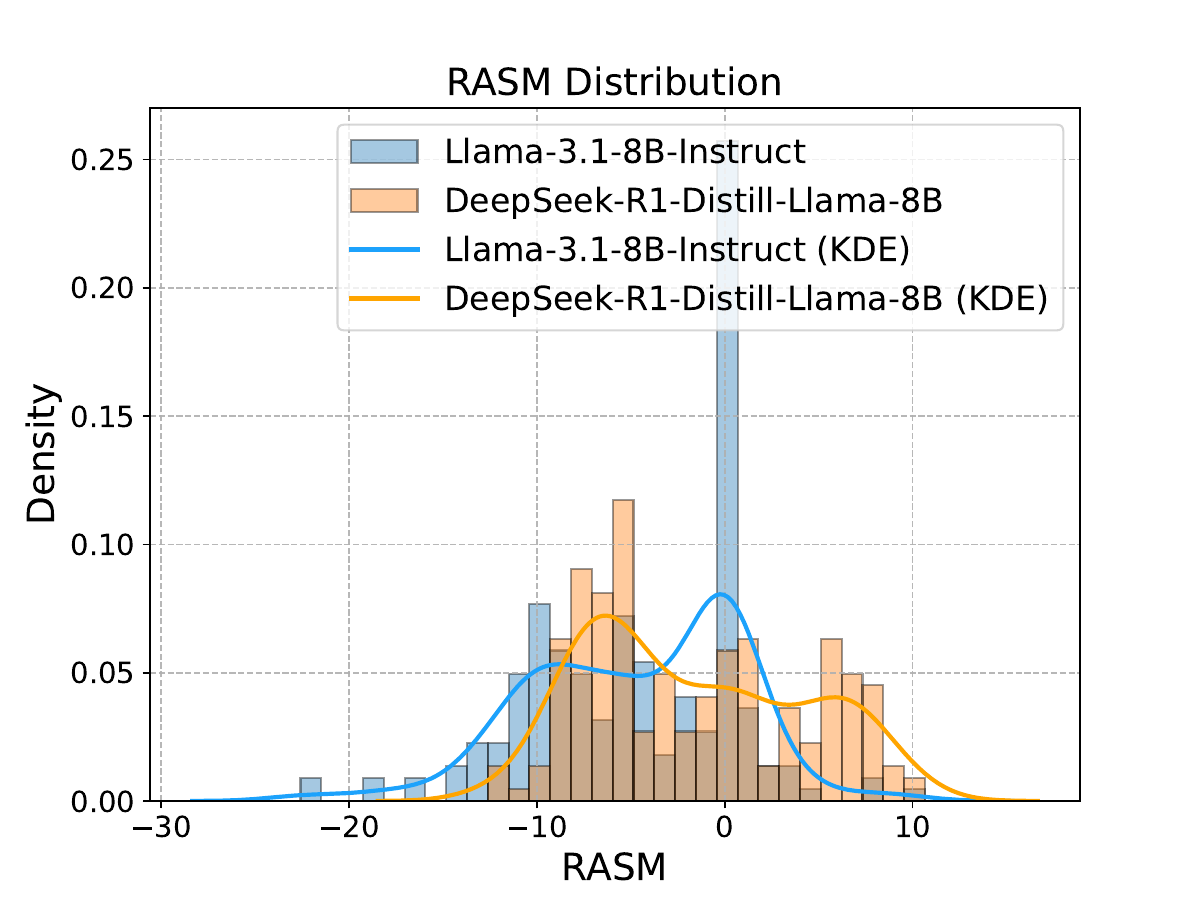} &
        \includegraphics[width=0.2\textwidth, trim=0.2cm 0.2cm 0.2cm 0.2cm, clip]{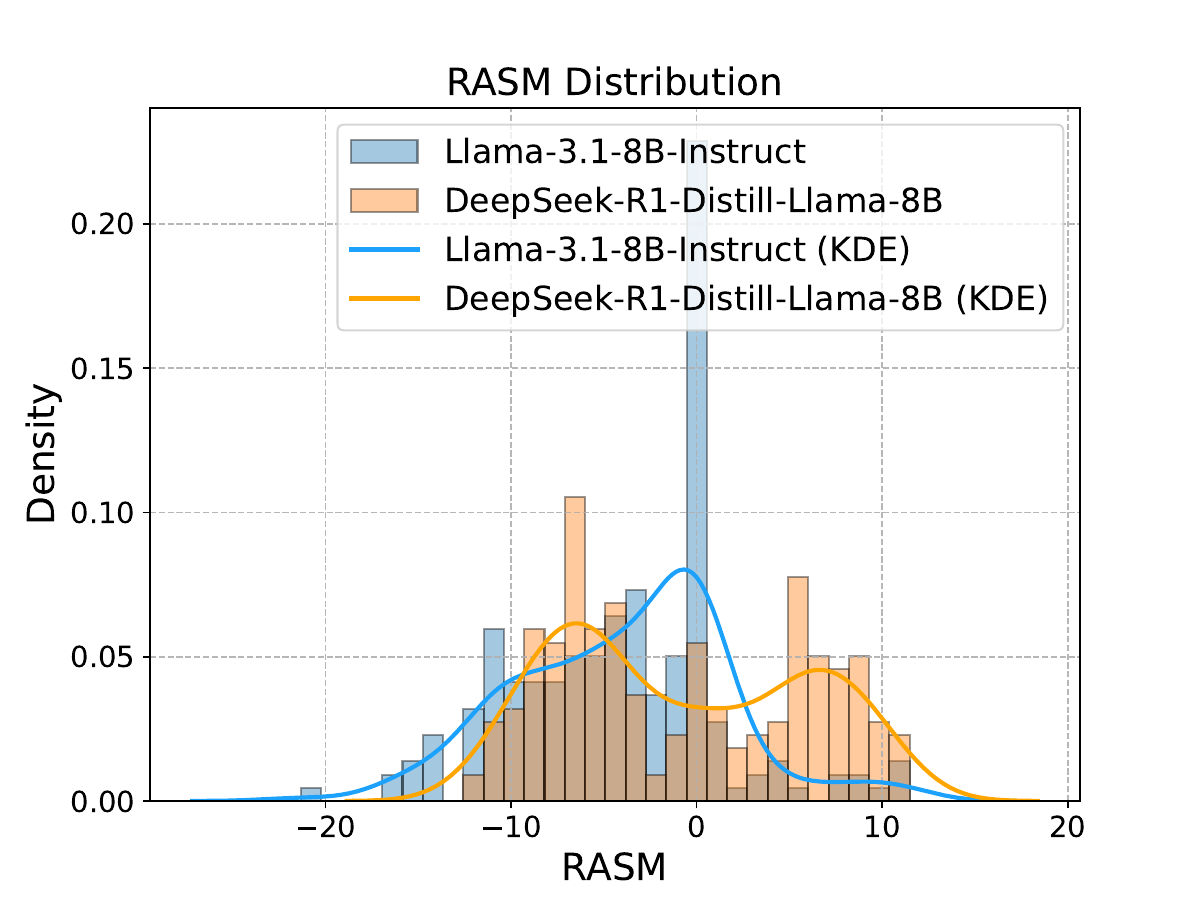} &
        \includegraphics[width=0.2\textwidth, trim=0.2cm 0.2cm 0.2cm 0.2cm, clip]{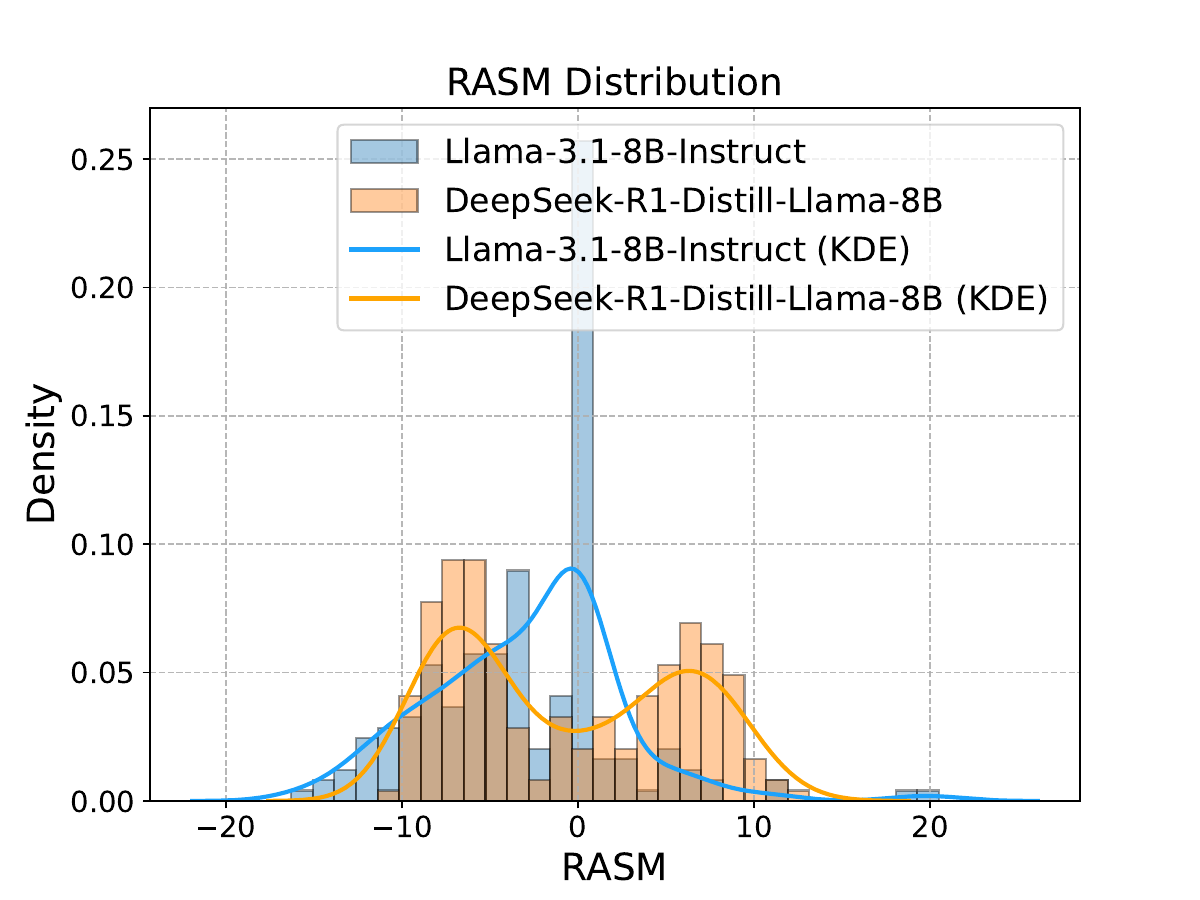} &
        \includegraphics[width=0.2\textwidth, trim=0.2cm 0.2cm 0.2cm 0.2cm, clip]{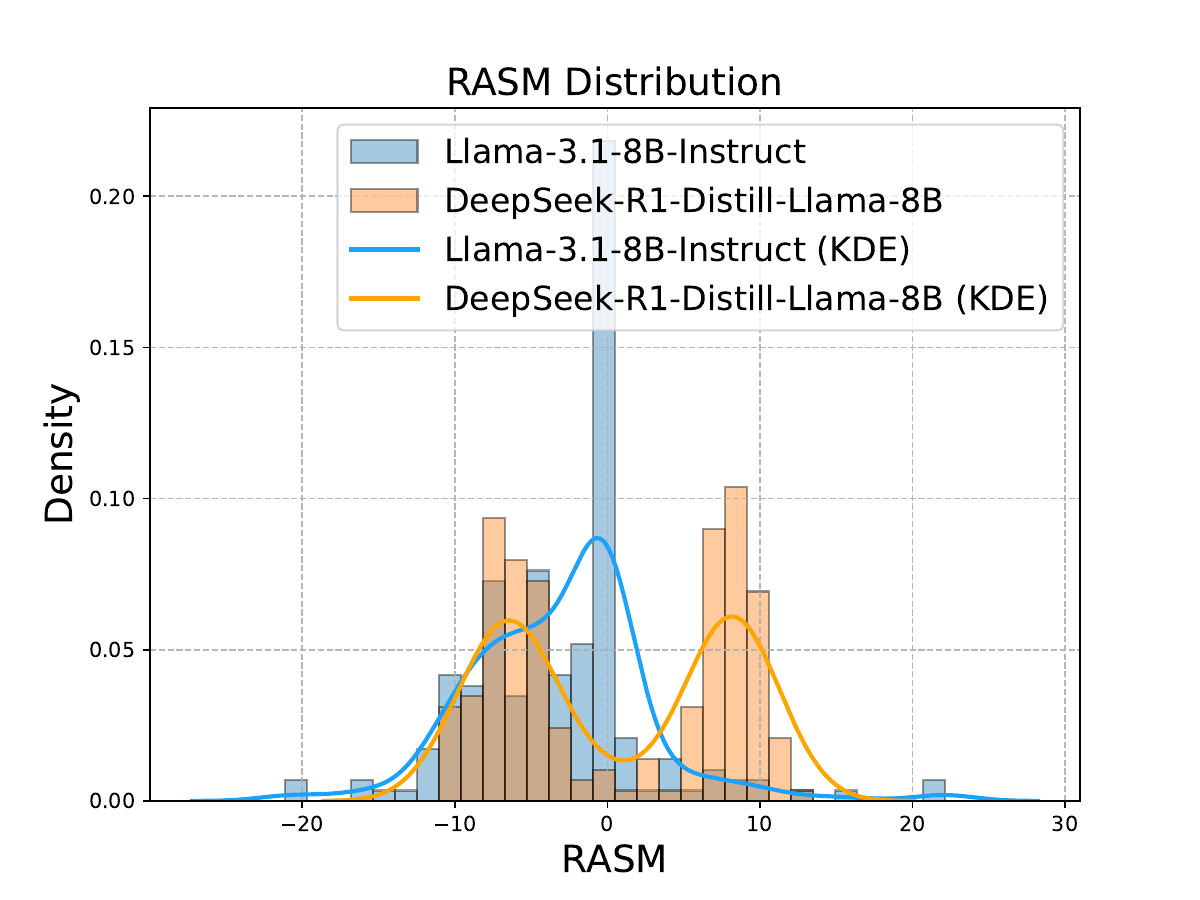} \\
        
        \includegraphics[width=0.2\textwidth, trim=0.2cm 0.2cm 0.2cm 0.2cm, clip]{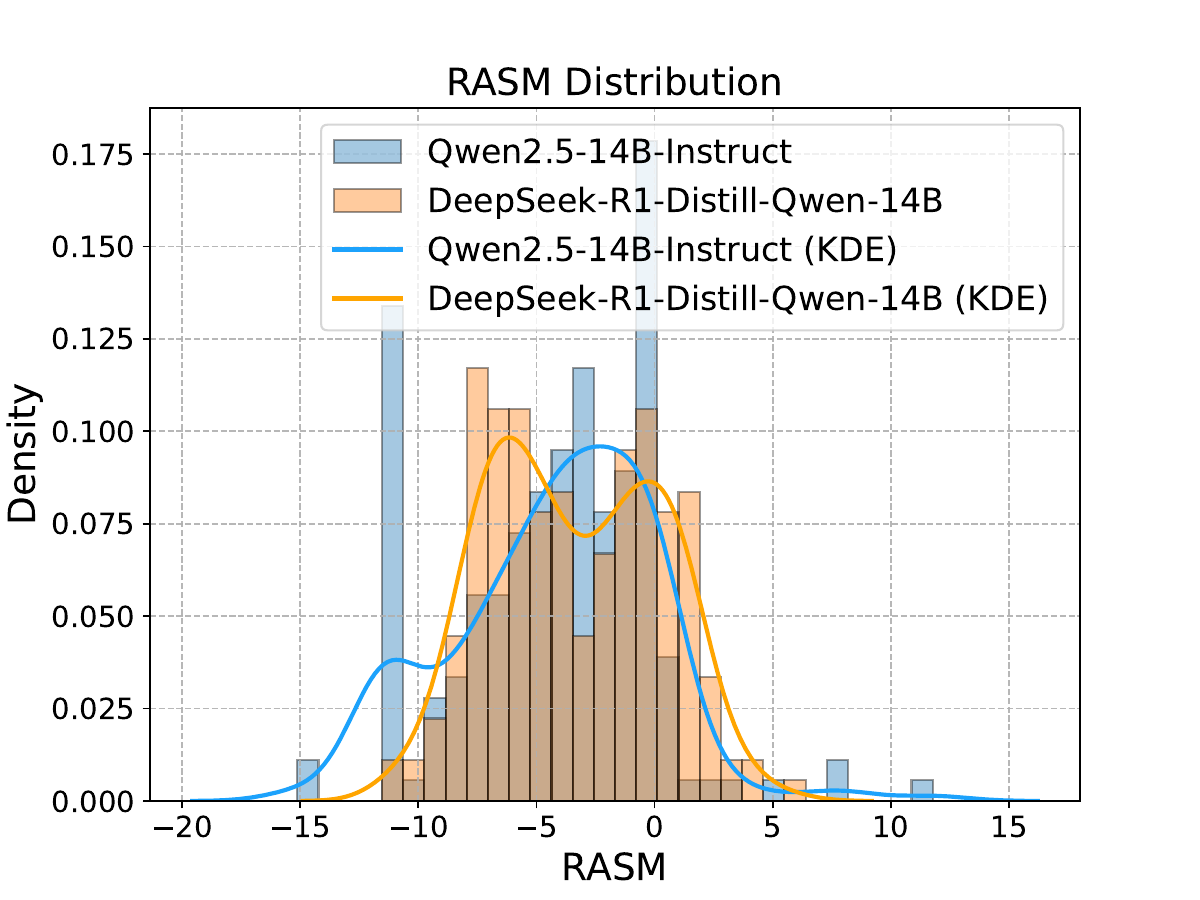} &
        \includegraphics[width=0.2\textwidth, trim=0.2cm 0.2cm 0.2cm 0.2cm, clip]{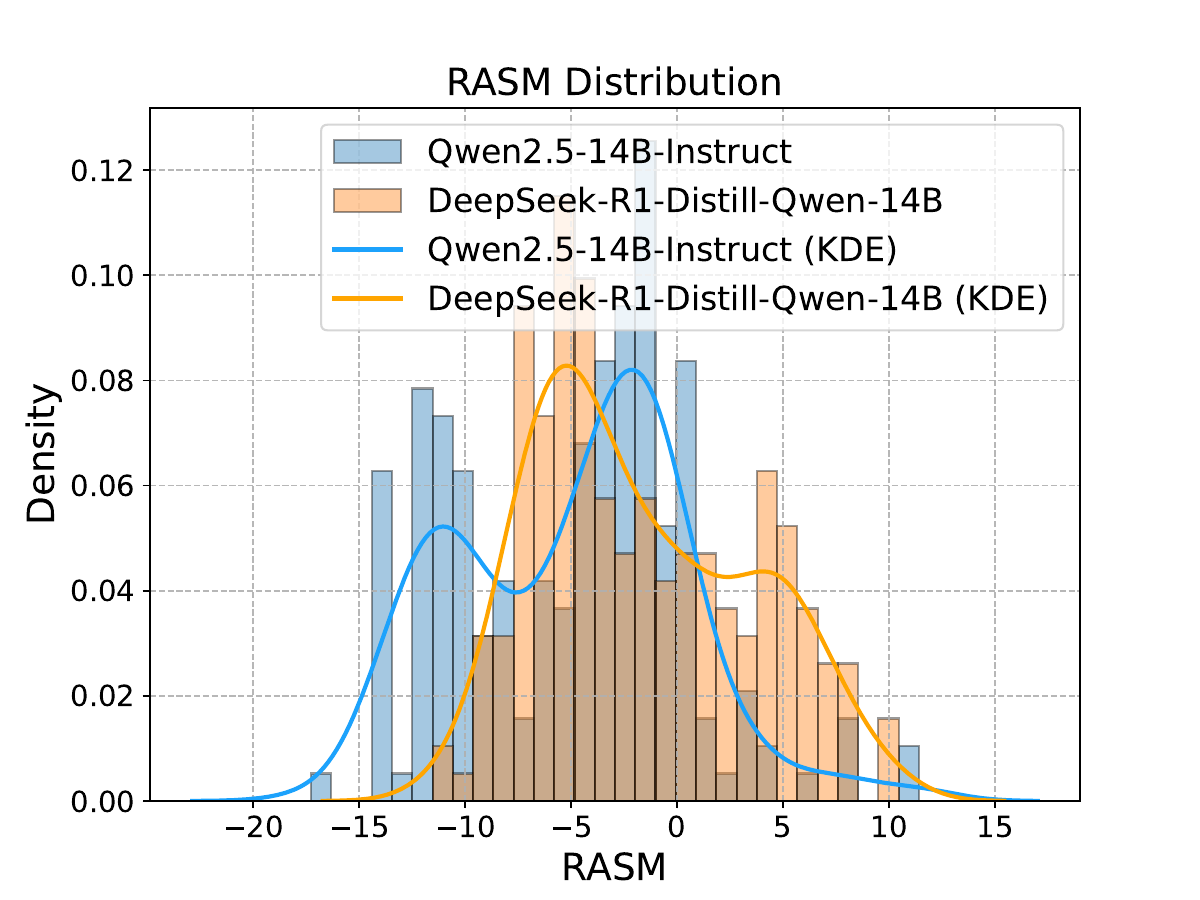} &
        \includegraphics[width=0.2\textwidth, trim=0.2cm 0.2cm 0.2cm 0.2cm, clip]{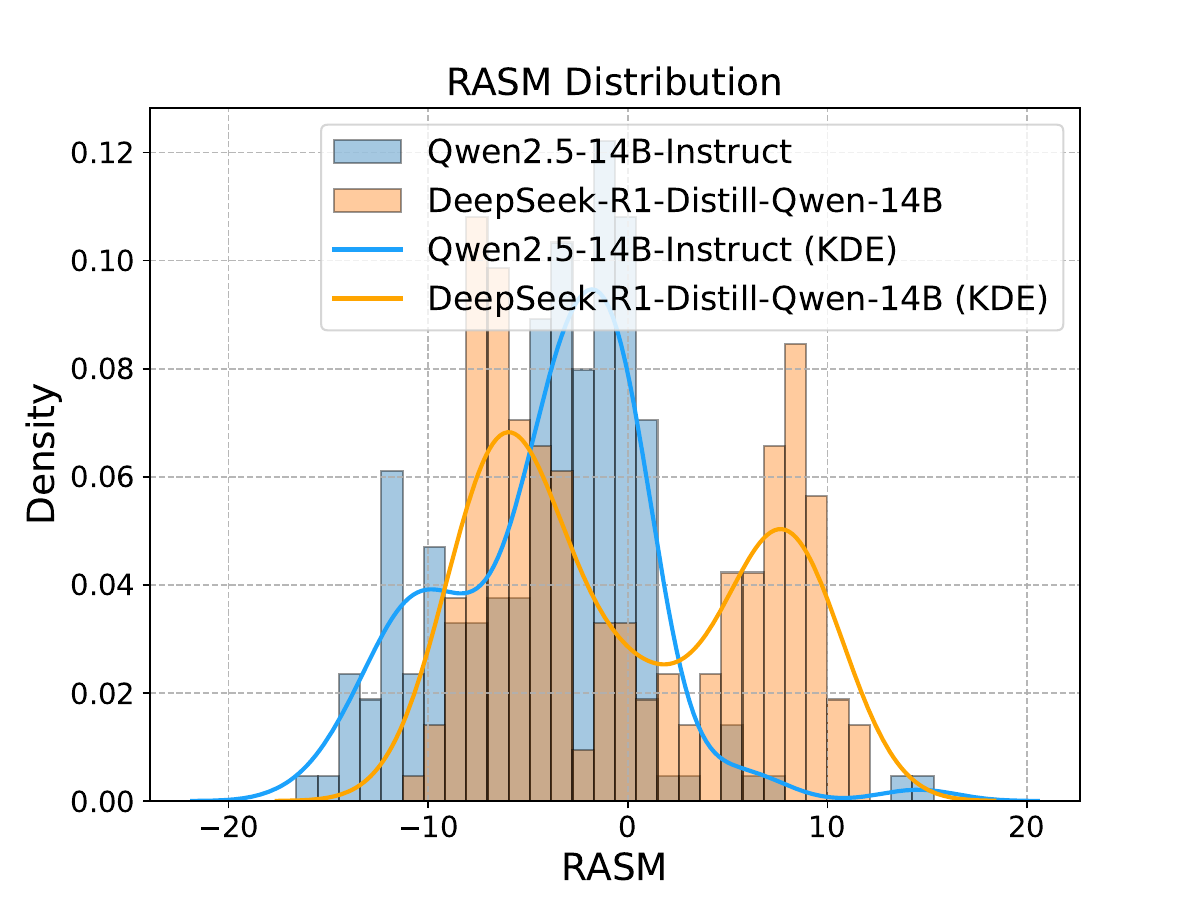} &
        \includegraphics[width=0.2\textwidth, trim=0.2cm 0.2cm 0.2cm 0.2cm, clip]{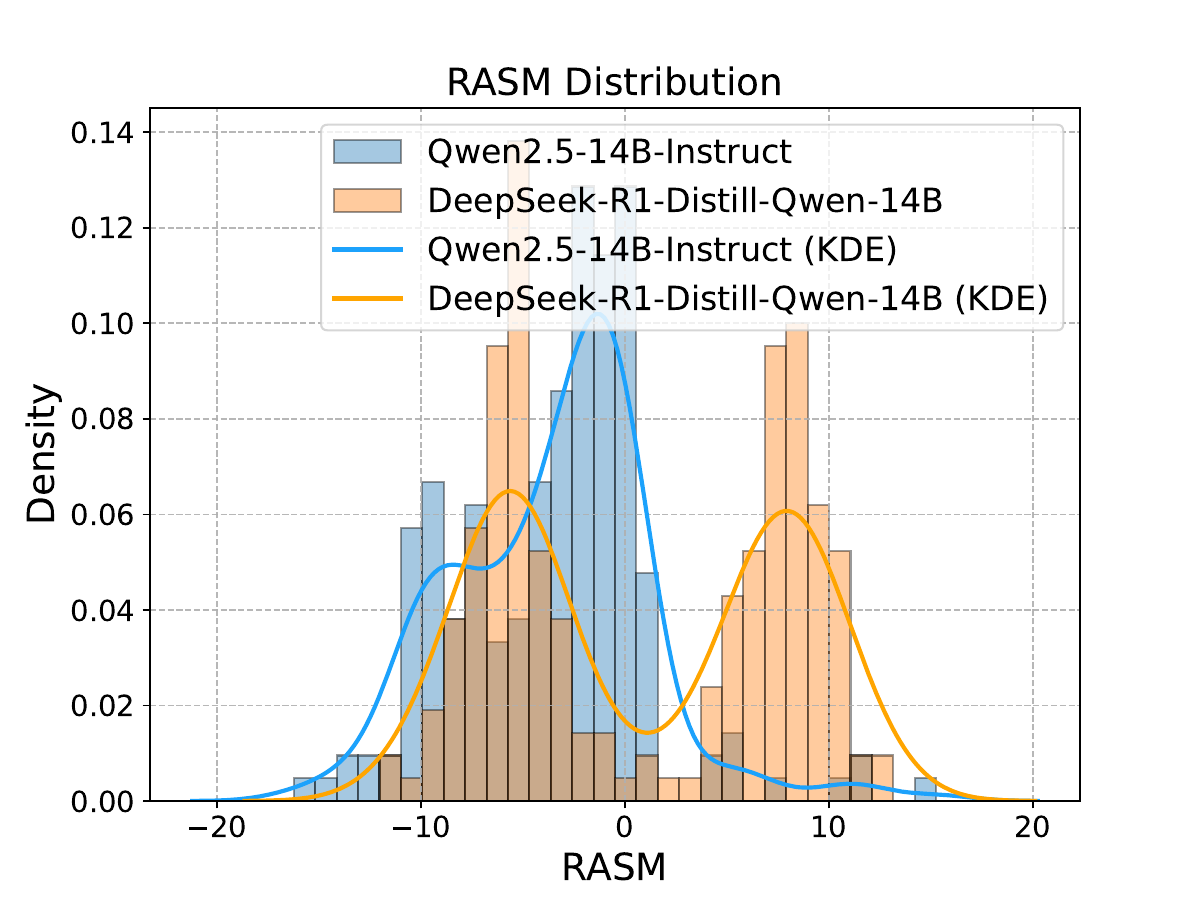} &
        \includegraphics[width=0.2\textwidth, trim=0.2cm 0.2cm 0.2cm 0.2cm, clip]{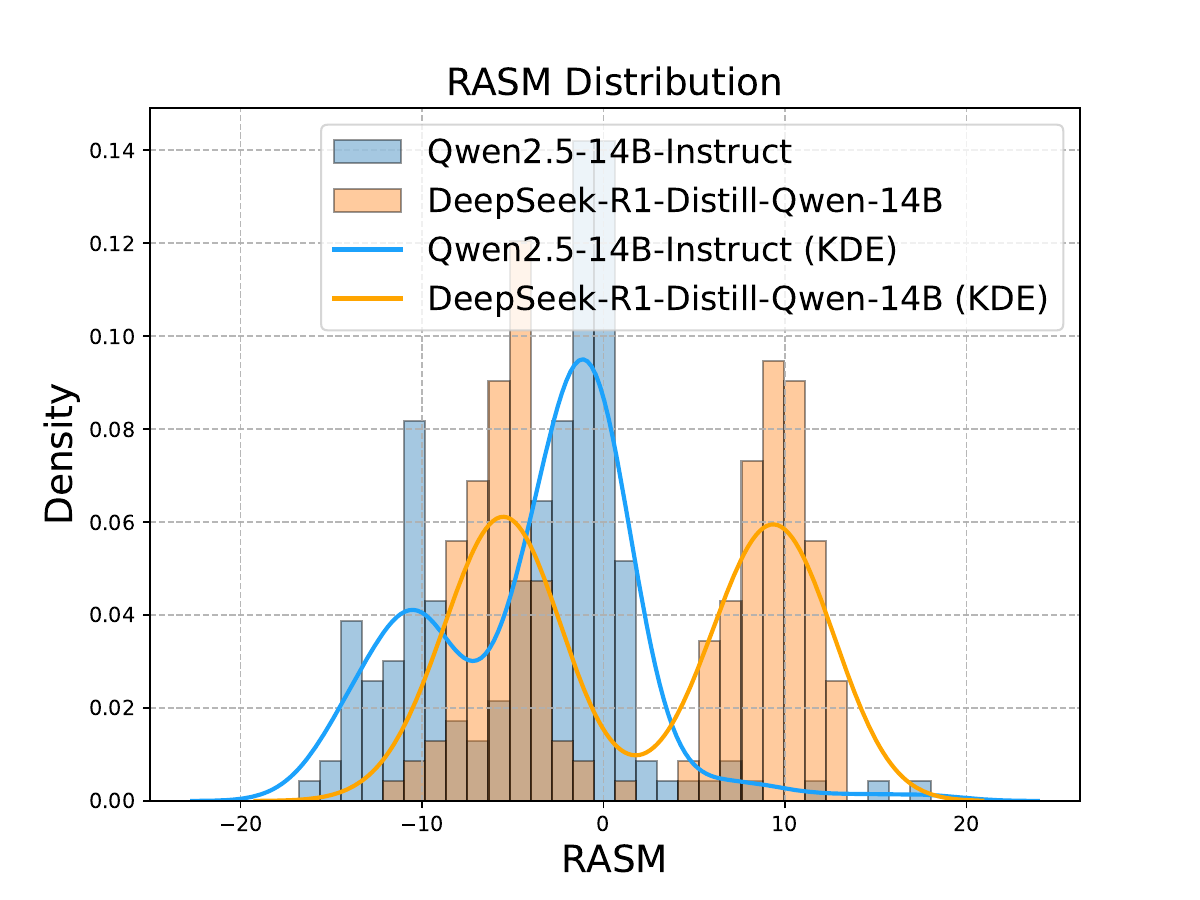} \\
        
        \includegraphics[width=0.2\textwidth, trim=0.2cm 0.2cm 0.2cm 0.2cm, clip]{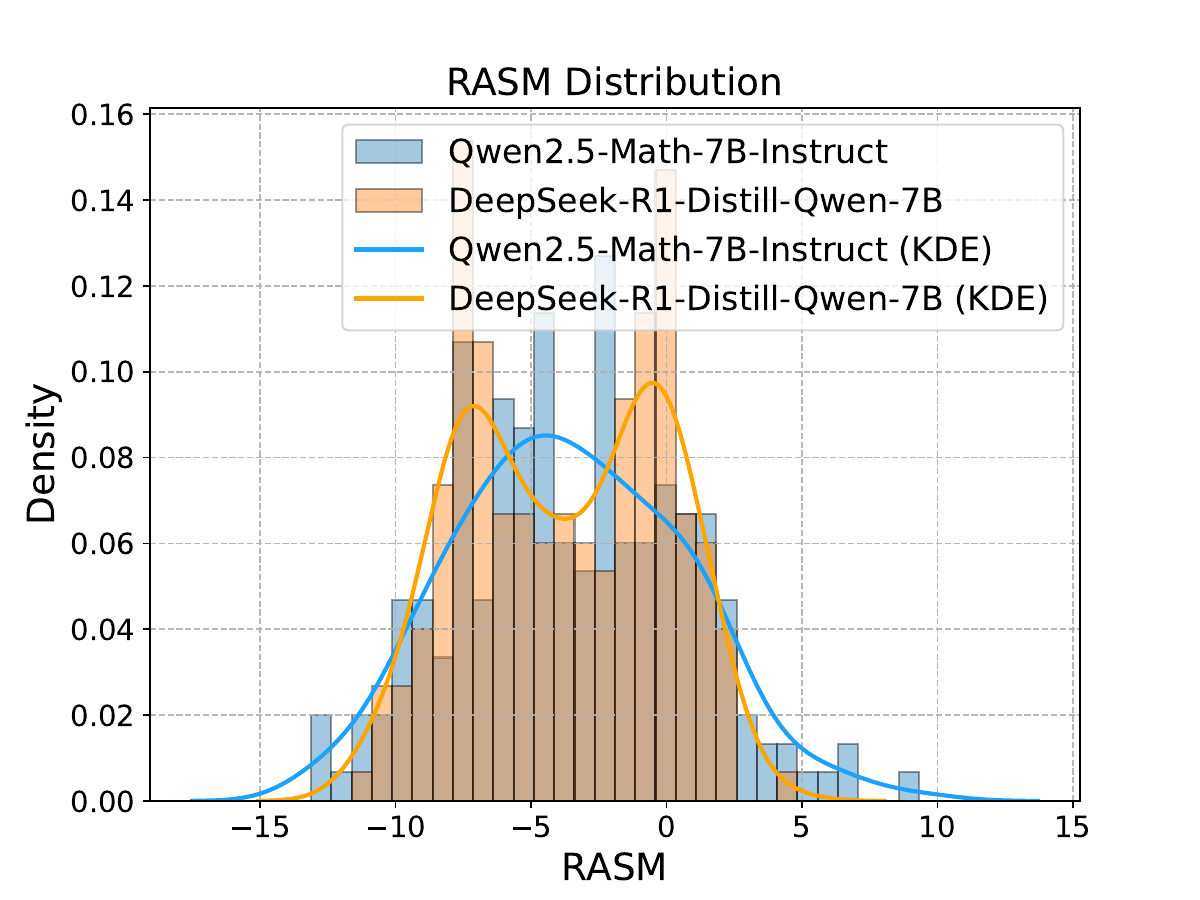} &
        \includegraphics[width=0.2\textwidth, trim=0.2cm 0.2cm 0.2cm 0.2cm, clip]{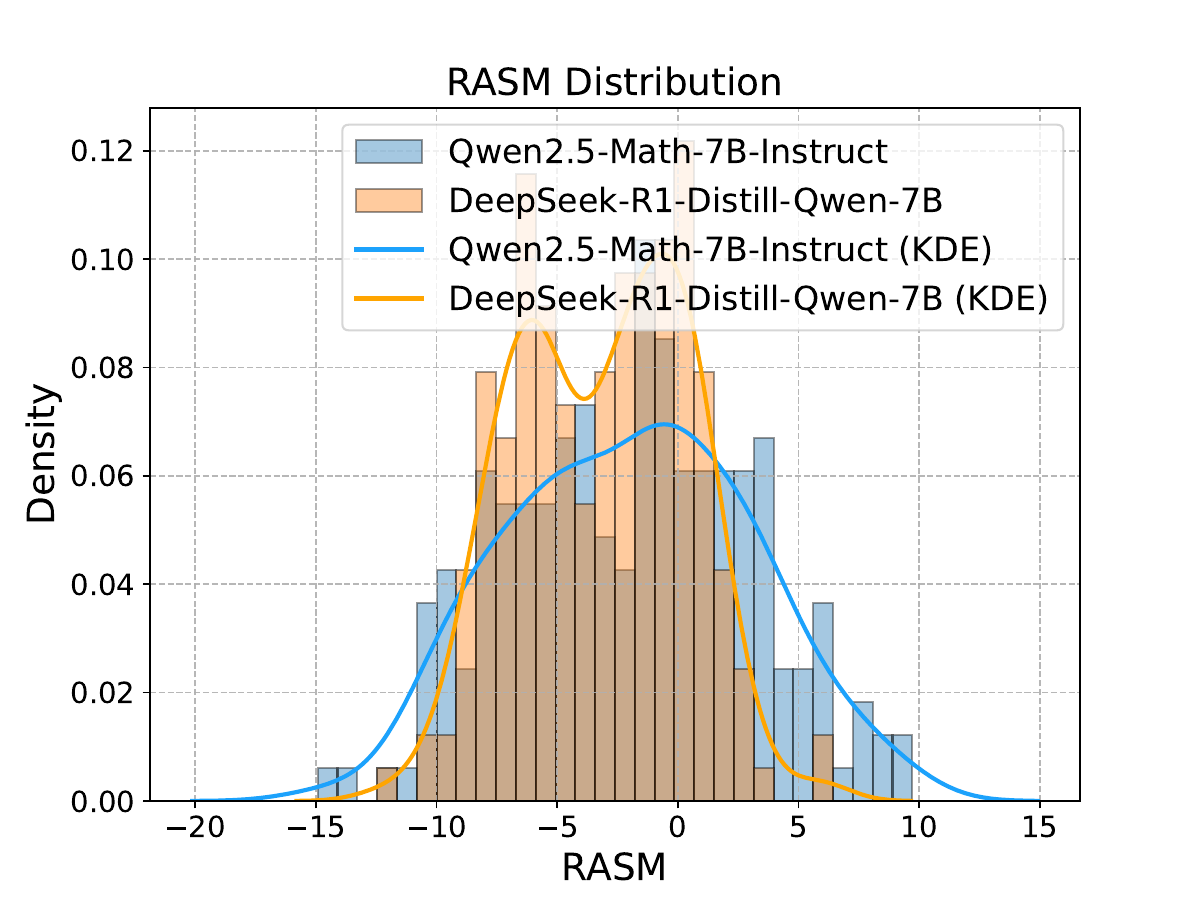} &
        \includegraphics[width=0.2\textwidth, trim=0.2cm 0.2cm 0.2cm 0.2cm, clip]{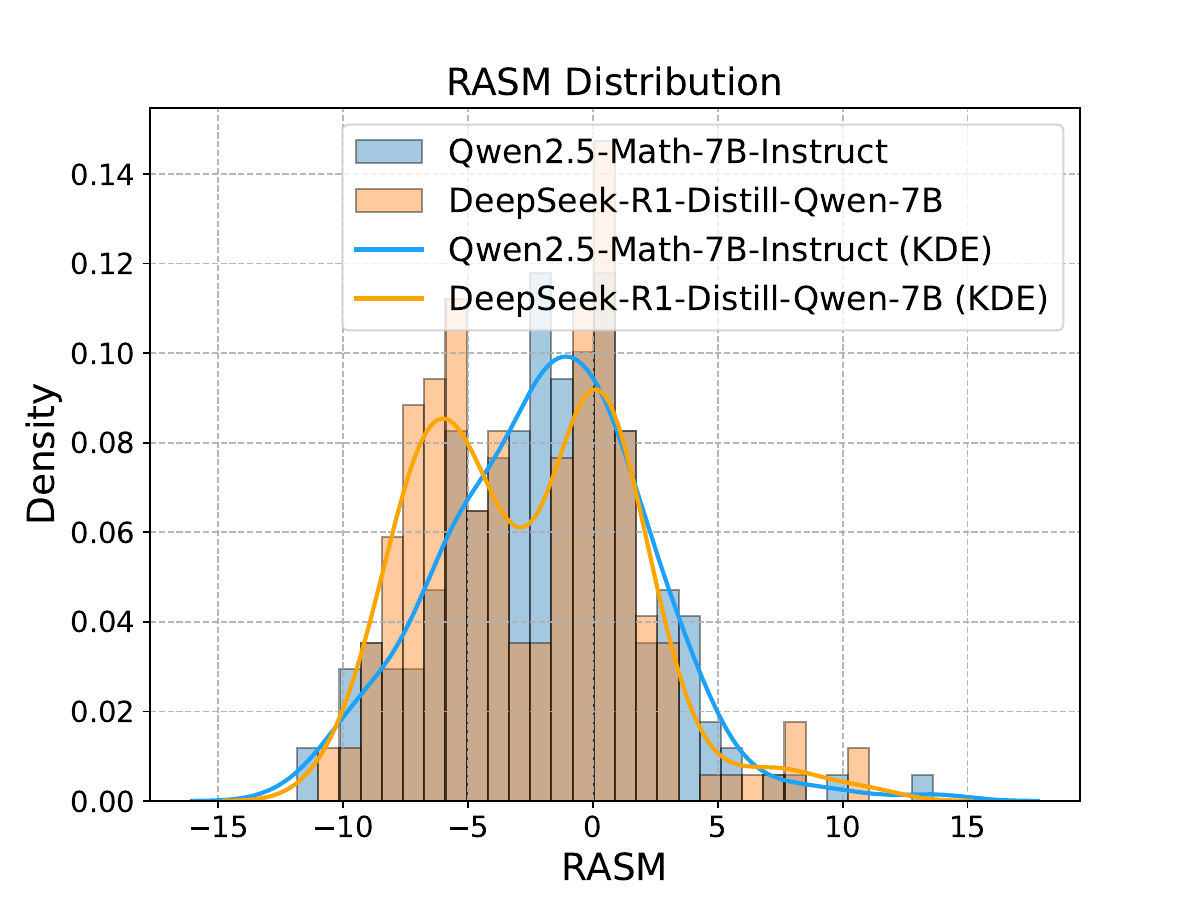} &
        \includegraphics[width=0.2\textwidth, trim=0.2cm 0.2cm 0.2cm 0.2cm, clip]{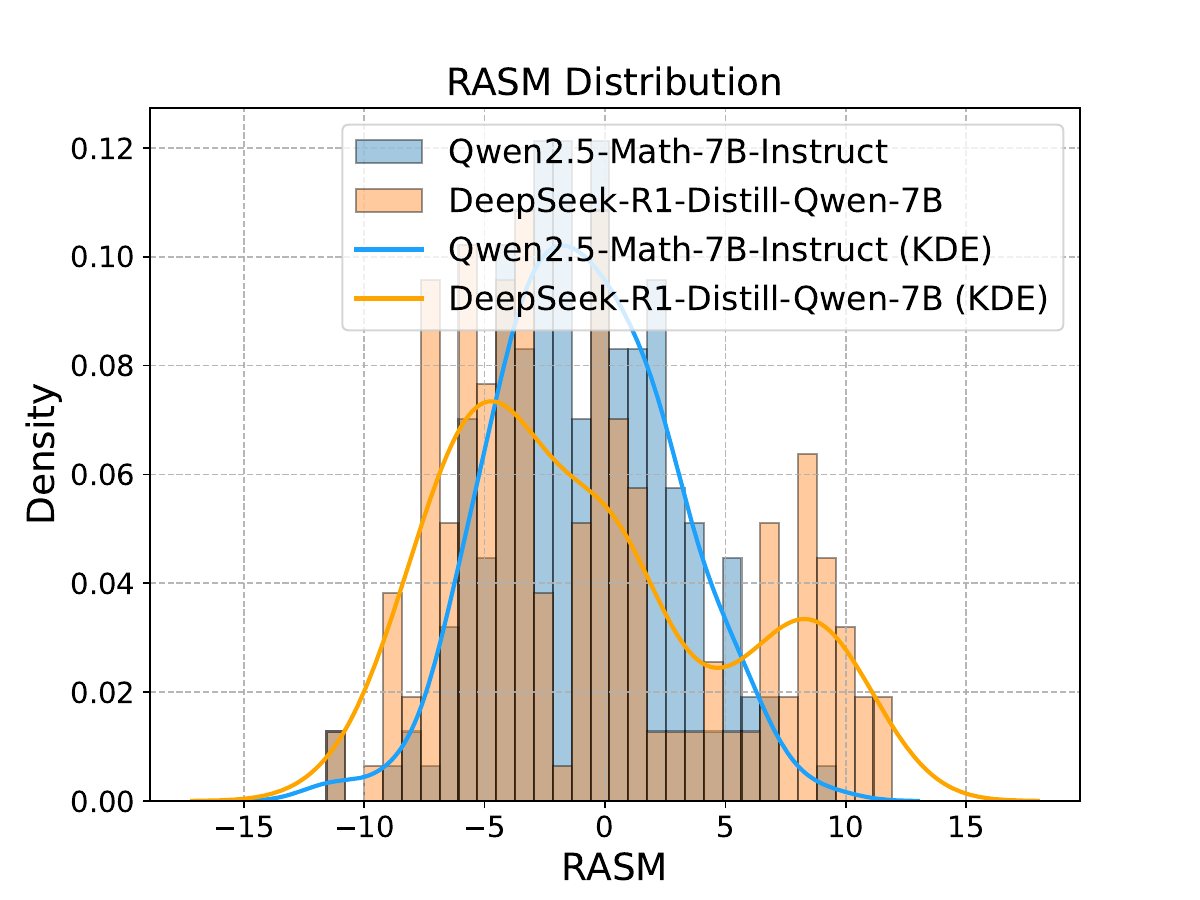} &
        \includegraphics[width=0.2\textwidth, trim=0.2cm 0.2cm 0.2cm 0.2cm, clip]{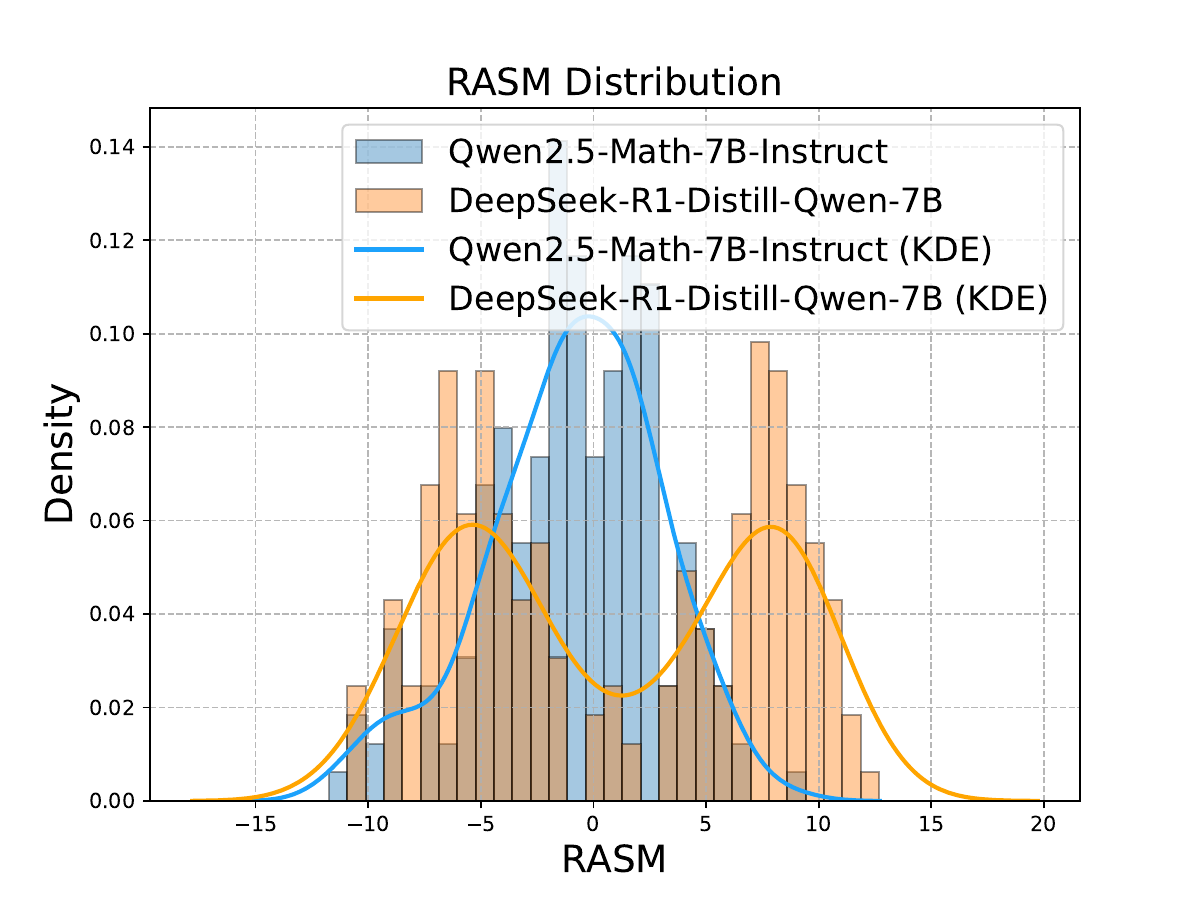} \\
        
        \includegraphics[width=0.2\textwidth]{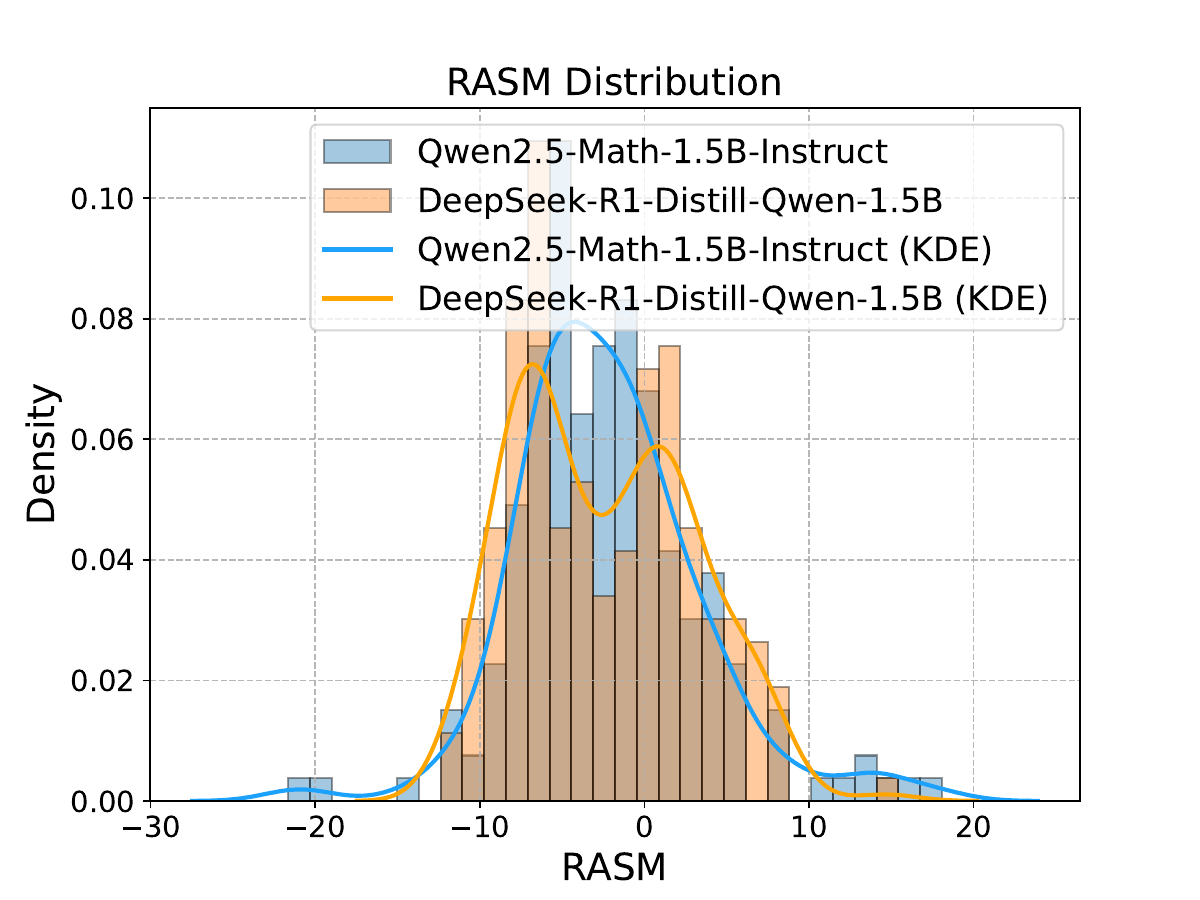} &
        \includegraphics[width=0.2\textwidth]{figure/rasm_results/4_DeepSeek-R1-Distill-Qwen-1.5B_vs_Qwen2.5-Math-1.5B-Instruct.pdf} &
        \includegraphics[width=0.2\textwidth]{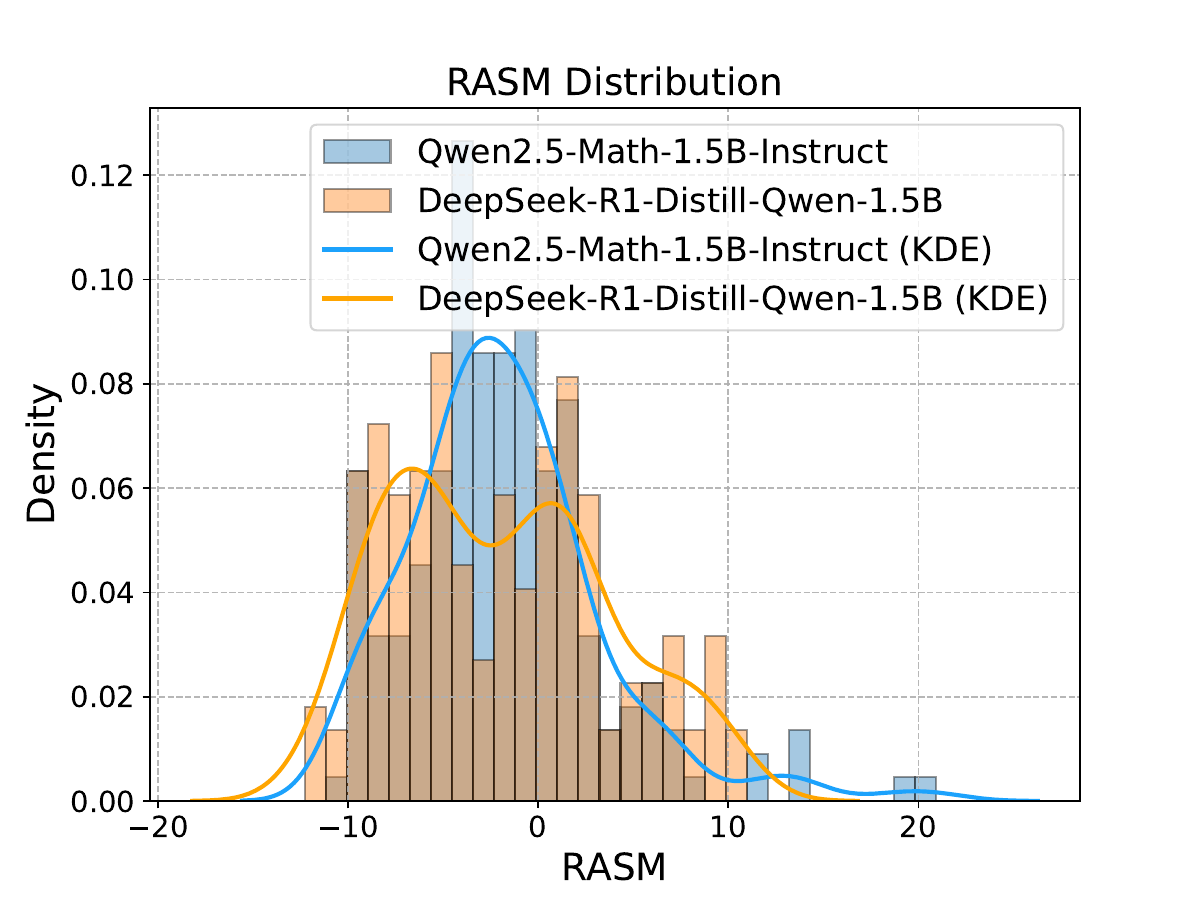} &
        \includegraphics[width=0.2\textwidth]{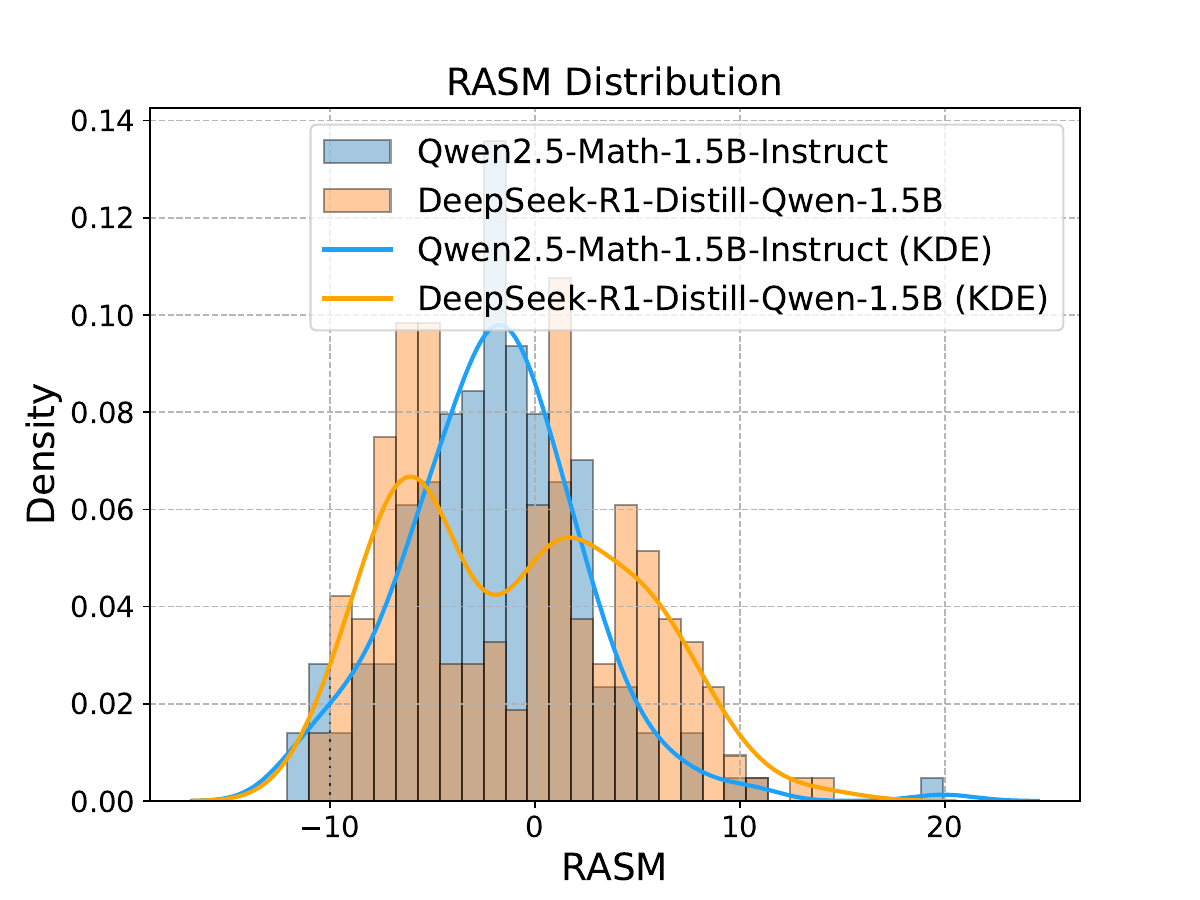} &
        \includegraphics[width=0.2\textwidth]{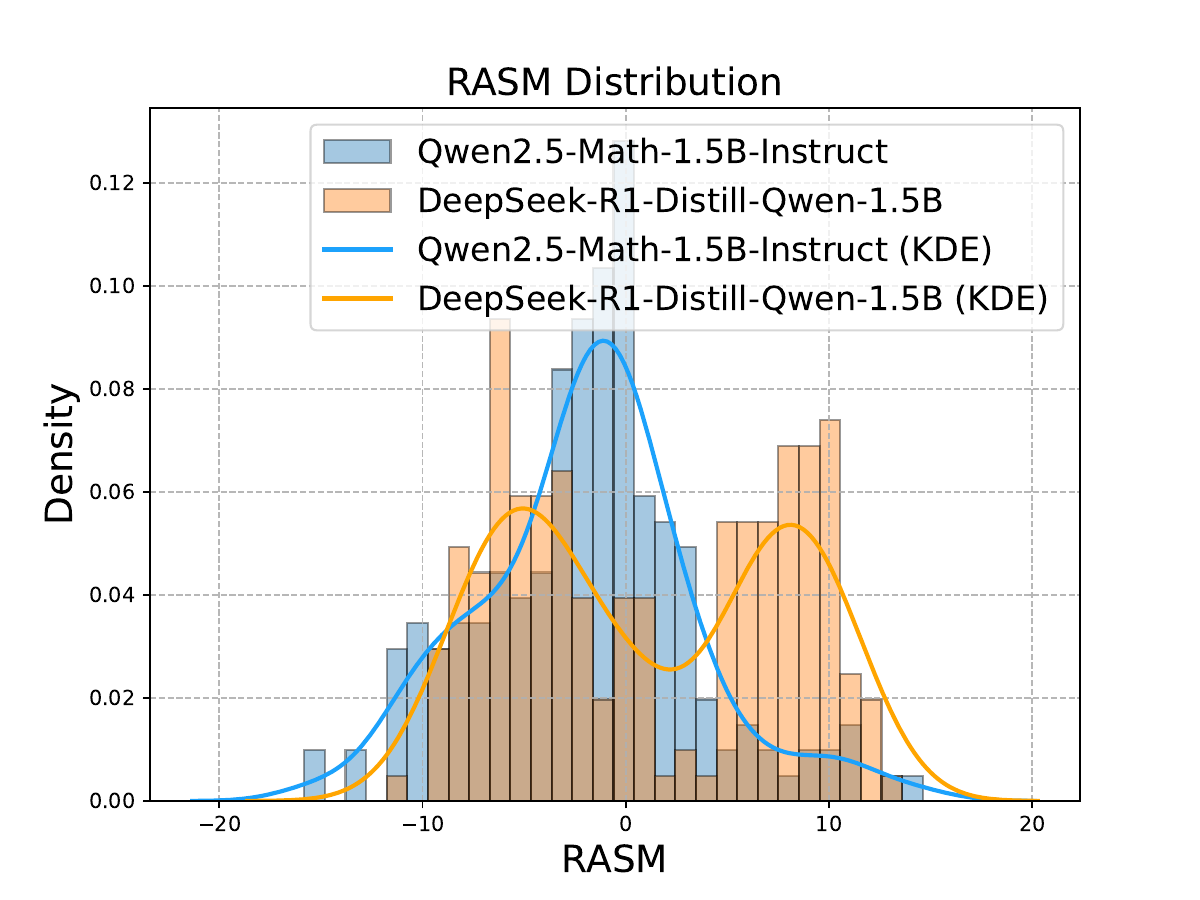} \\ 
    
        \hline
    \end{tabular}
\end{adjustbox}
\caption{Comparison of RASM across different difficulty levels $n$ between \textcolor{myorange}{aha} and \textcolor{myblue}{no-aha} model pairs. We fit the score distributions for both models using Kernel Density Estimation (KDE) with a bandwidth selection method.}
\label{fig:rasm-comparison}
\vspace{-0.2cm}
\end{figure}
In \S~\ref{subsec:linguistic pattern}, we observed that aha models consistently utilize a small set of anthropomorphic tokens (e.g., ``Wait'' and ``Hmm'') to start their self-reflection, search, and verification behaviors. This observation enables us to efficiently evaluate a model's expressive changes by examining its next-token probability distribution, rather than relying on external evaluators as in \citet{cheng2024anthroscore} or LLM-as-evaluator approaches \citep{gu2024survey}. To evaluate the change formally for each model, we manually identify a set of anthropomorphic beginning tokens, denoted as $\mathcal{A}$, selected from tokens most likely to signal the aha model's shift to an anthropomorphic tone. Additionally, we define a set of pure reasoning beginning tokens like ``Calculate'', ``First'', $\mathcal{R}$, which models frequently use when performing direct reasoning or computational steps. The selection principle is based on our results in \S~\ref{subsec:linguistic pattern} and is detailed in Appendix~\ref{app:tokenselection}. We can then estimate the probability of anthropomorphic versus pure reasoning responses as the sum of the probabilities at the beginning of a sentence assigned to tokens in $\mathcal{A}$ and $\mathcal{R}$, respectively. Given a vector of next-token probabilities $\textbf{p} = \{p_1,p_2, \dots, p_{|\mathcal{V}|}\} \in \mathbb{R}^{|\mathcal{V}|}$, where $\mathcal{V}$ denotes the model's vocabulary, we define $P_\text{A}(\textbf{p}) = \sum_{t \in \mathcal{A}}p_t$ and $P_\text{R}(\textbf{p}) = \sum_{t \in \mathcal{R}}p_t$. We calculate the log odds ratio between these probabilities to form our Reasoning-Anthropomorphic Separation Metric (RASM):
$$\text{RASM}(\textbf{p}) = \log\left(\frac{P_{\text{A}}(\textbf{p}) + \epsilon}{P_{\text{R}}(\textbf{p}) + \epsilon}\right) = \log\left(\sum_{t \in \mathcal{A}} p_t + \epsilon \right) - \log\left(\sum_{t \in \mathcal{R}} p_t + \epsilon \right),$$
where $\epsilon = 1\mathrm{e}{-10}$ is added for numerical stability. 

We run all the models listed in Table~\ref{tab:models} on the K\&K and Count Down test sets across different difficulty levels, computed the RASM at the beginning position of each sentence in the responses, and then displayed the RASM density for all models in Figure~\ref{fig:rasm-comparison}. Additionally, we include the statistical distribution measures such as the Mean, Skewness, Kurtosis, and Bimodality Coefficient in Appendix~\ref{app:Additional results for Reasoning-Anthropomorphic Separation}, Figure~\ref{fig: Distribution Metrics}. 
We observe that both aha and no-aha models show an increasing trend in average RASM as task difficulty rises. However, Figure~\ref{fig: Distribution Metrics} shows that no-aha models consistently maintain a mean RASM below zero, indicating their preference for pure reasoning. In contrast, aha models exhibit a sharper rise, eventually surpassing zero, reflecting a stronger shift toward anthropomorphic expression.
Also, the aha models exhibit a stronger bimodal distribution, can be evidenced by the rising bimodality coefficient in Figure~\ref{anthropomorphic_and_reasoning_tokens}.
This phenomenon suggests a sophisticated linguistic mechanism where the models develop a clearer separation between reasoning and anthropomorphic self-reflection compared to no-aha model. 
As task difficulty increases, the aha models gradually symmetrize their linguistic approach, balancing anthropomorphic insights with pure reasoning. Notably, their initially right-skewed distribution shifts toward zero, indicating that the models are not merely adding anthropomorphic elements arbitrarily, but are strategically integrating them to enhance problem-solving capabilities.

This finding suggests that the emergence of the "aha moment" is more than a stylistic linguistic pattern; it signifies a deeper refinement of reasoning ability. By fostering a distinct separation between anthropomorphic and reasoning features, aha models enable a more adaptive and flexible problem-solving framework, particularly as task complexity increases.


\subsection{Contrastive Analysis in Latent Space}
\label{subsec:contrastiveanalysis}
\begin{figure}[h]
    \centering
    \includegraphics[width=\linewidth, trim=0 0 0 0, clip]{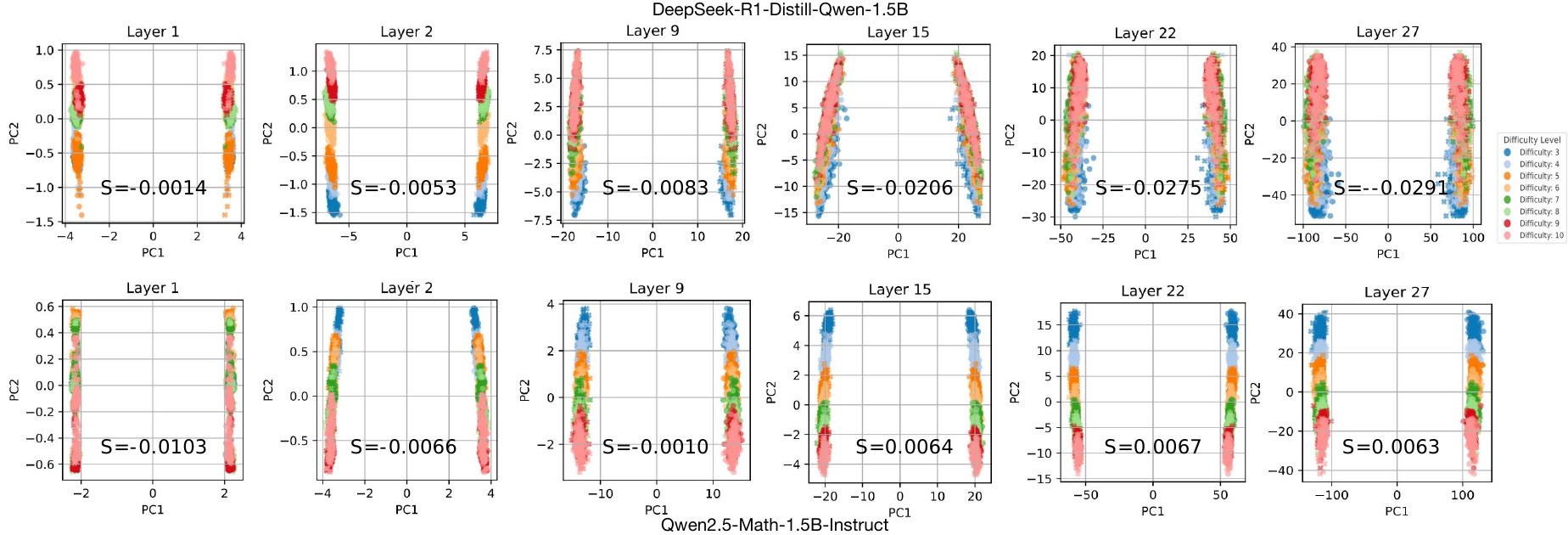}
    \includegraphics[width=\linewidth, trim=0 0 0 0, clip]{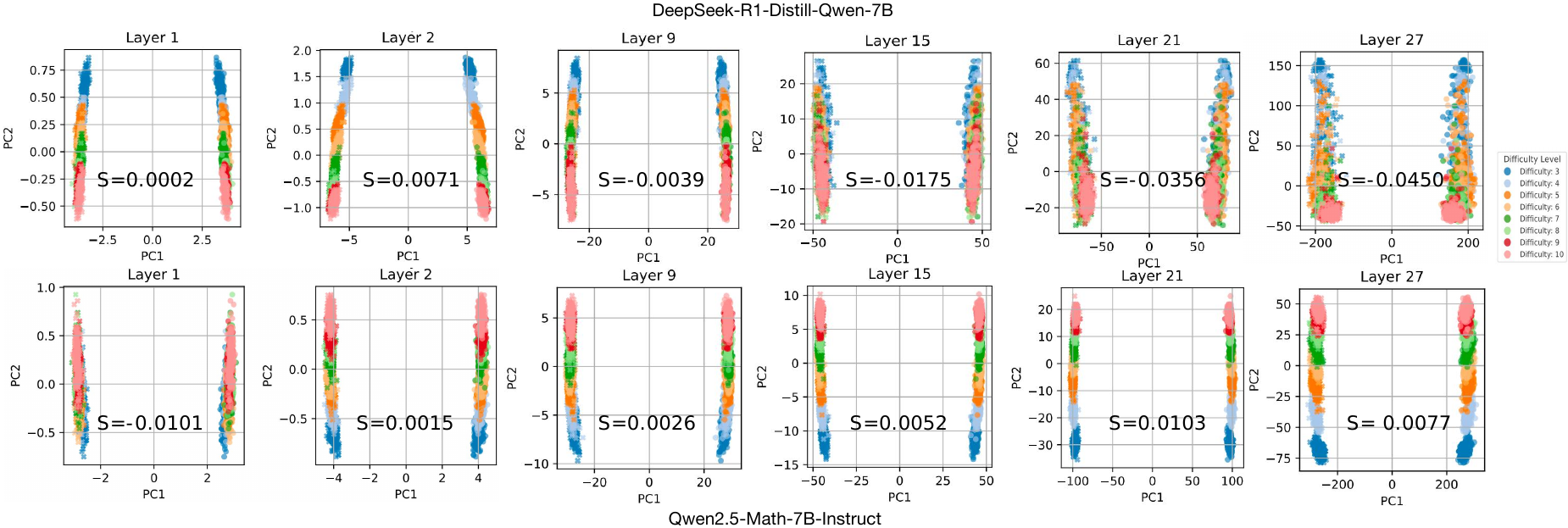}
    \caption{Difficulty level clustering of different models, with $S$ representing the silhouette score~\citep{rousseeuw1987silhouettes} for clustering based on the second principal component (PC2). A higher $S$ indicates better clustering and feature separation performance. 
    }
    \label{fig:feature_split_crosstask_temp}
    \vspace{-0.3cm}
\end{figure}

To further explore how the ``aha moment'' influences the way models handle complex reasoning problems of varying difficulty, this section analyzes the behavior of the aha and no-aha models in the latent space of each layer.

For each model $\mathcal{M}$ in Table~\ref{tab:models}, we run them on the K\&K and Count Down training datasets and project the activations of its $l$-th layer corresponding at the last token position for each difficulty level using PCA, inspired by \citet{rimsky-etal-2024-steering}. While it is evident that activations can always be separated across different tasks, our goal is to investigate further how these models encode features at varying difficulty levels, building on previous findings.  We surprisingly find that, unlike the no-aha models such as Qwen2.5-Math-1.5B-Instruct and Qwen2.5-Math-7B-Instruct (Figure~\ref{fig:feature_split_crosstask_temp}), where difficulty clustering becomes progressively clearer as the layer depth increases, the aha models, including DeepSeek-R1-Distill-Qwen-1.5B and DeepSeek-R1-Distill-Qwen-7B, exhibit a well-defined difficulty clustering pattern in the early layers. However, as layer depth increases, this clustering pattern becomes increasingly blurred. To quantify this observation, we use the silhouette score ($S$)~\citep{rousseeuw1987silhouettes} as a measure of this clustering quality. In the no-aha models, the silhouette score gradually improves across layers, starting from $S = -0.0103$ (Qwen2.5-Math-1.5B-Instruct) and $S = -0.0101$ (Qwen2.5-Math-7B-Instruct), reaching their respective peaks at layer 22 with $S = 0.0067$ and $S = 0.0103$, indicating an increasing separation of difficulty levels. In contrast, for the aha models, the silhouette score declines with depth, suggesting a loss of cluster separability. DeepSeek-R1-Distill-Qwen-1.5B starts at $S = -0.0053$(layer 2) and decreases to $S = -0.0291$ (layer 27), while DeepSeek-R1-Distill-Qwen-7B drops from $S = 0.0071$ (layer 2) to $S = -0.045$ (layer 27). This trend suggests that while early layers in aha models already encode distinct difficulty-related representations, deeper layers increasingly mix them, blurring the clustering. 

Furthermore, we demonstrate that the no-aha model exhibits a more symmetrical separation of difficulty features, as seen in Figure~\ref{fig:feature_split_crosstask_temp}. The PC2 projections in no-aha models remain largely symmetric across layers, with a balanced distribution of data points along the positive and negative ranges of the PC2 axis. In contrast, for the aha model, the scale of the projections in the simple difficulty direction gradually surpasses that of the hard direction as the model depth increases. For instance, in DeepSeek-R1-Distill-Qwen-7B at layer 27, the positive range of the PC2 axis is significantly larger than the negative range, and more difficult problem samples are projected in the same direction as simpler problems, compared to Qwen2.5-Math-7B-Instruct. Additionally, for easier problems—such as those at difficulty levels 3–4—the aha model projects them across a wider range rather than strictly separating them within a limited space, as seen in no-aha models. This suggests that the aha model may inherently treat simpler problems as requiring more complex processing, potentially incorporating more nuanced reasoning even at lower difficulty levels.

These findings indicate that the ``aha moment'' influences the model’s ability to recognize the difficulty of a problem in its early layers. Then it appears to help the model learn to approach simpler problems with greater complexity while trying to simplify harder problems to support reasoning, potentially mitigating issues such as the Reasoning Collapse discussed in \S~\ref{subsec:Reasoning collapse}

\vspace{-0.2cm}
\section{Conclusion}
\vspace{-0.1cm}
In this paper, we conduct a comprehensive study of the ``aha moment'' to explore how it helps the model perform complex reasoning. We demonstrate that the outputs of the ``aha model'' exhibit more anthropomorphic characteristics, and as the difficulty increases, these anthropomorphic traits become more frequent. Moreover, it is closely tied to the model's expression of its uncertainty. Furthermore, we find that the probability of ``reasoning collapse'' in the ``aha model'' significantly decreases when answering complex problems, suggesting that the ``aha moment'' helps the model complete reasoning successfully rather than getting stuck in repetition or reasoning loops. Additionally, by analyzing the model's latent space, we discover that, unlike the ``no-aha model,'' which progressively deepens its understanding of problem difficulty layer by layer, the ``aha model'' can effectively recognize the problem's difficulty at earlier layers. However, it tends to blur the boundaries of difficulty in later layers, offering an explanation for the ``overthinking'' issue in LRMs.

\newpage
\bibliography{colm2025_conference}
\bibliographystyle{colm2025_conference}

\appendix
\newpage
\section{Further Experiment Details}

\subsection{Chat templates}
The chat templates used in all experiments are presented in Table~\ref{tab:chattemplate}, and all are aligned with the default settings of the Hugging Face generation.

\begin{table*}[!ht]
\centering
\renewcommand\tabcolsep{0.9pt}
\renewcommand{\arraystretch}{1.5}
\resizebox{\textwidth}{!}{
\small
\begin{tabular}{@{}p{3.5cm} p{11cm}}
\toprule
\textbf{Model Family}&\textbf{Corresponding Chat Templates} \\ \midrule
DeepSeek-R1-Distill & {\texttt{"<|begin\_of\_sentence|><|User|>\color{myskyblue}\{QUERY\}\color{myred}<|Assistant|><think>\string\n"}}\\
Qwen2.5-Math-Instruct& {\texttt{"<|im\_start|>user\string\n\color{myskyblue}\{QUERY\}\color{myred}<|im\_end|>\string\n<|im\_start|>assistant\string\n\color{black}"}}\\
Llama-3.1-Instruct& 
\texttt{"<|begin\_of\_text|><|start\_header\_id|>system<|end\_header\_id|>\string\n\string\n Cutting Knowledge Date: December 2023 \string\n Today Date: 26 Jul 2024\string\n \string\n<|eot\_id|><|start\_header\_id|>user<|end\_header\_id|>\string\n\string\n \textcolor{myskyblue}{\{QUERY\}} \color{myred}<|eot\_id|><|start\_header\_id|>assistant<|end\_header\_id|>\string\n\string\n\color{black}"}
\\ 
\bottomrule
\end{tabular}
}

\caption{Model families and their corresponding chat templates. The term \textcolor{myskyblue}{\texttt{QUERY}} refers to the question or puzzle provided to the model to solve, while \textcolor{myred}{post-query sequence} denotes the sequence that follows the query, as applied by the model's chat template.}

\label{tab:chattemplate}
\end{table*}

\subsection{Reasoning-Anthropomorphic token selecction}
\label{app:tokenselection}
\paragraph{Aha token and no-aha token separation strategy.}
To systematically examine the differences in capabilities between the two models, we segmented the output texts into sentences and analyzed the first token of each sentence. We defined the initial token in the output of the no-aha model as the ``\textcolor{myblue}{no-aha token}'' and the initial token in the output of the aha model as the ``\textcolor{myorange}{aha token}''. 

To further separate the two types of tokens, we calculate and rank the contrast probability for each token. The contrast probability \( P_{\text{contrast}}(t) \) is defined as follows:

For aha tokens:
\[
P_{\text{contrast}}(t) = P_{\text{aha}}(t) - P_{\text{no-aha}}(t)
\]
For no-aha tokens:
\[
P_{\text{contrast}}(t) = P_{\text{no-aha}}(t) - P_{\text{aha}}(t)
\]

Where:
\begin{itemize}
    \item \( P_{\text{aha}}(t) \) is the frequency proportion of token \( t \) among all aha tokens;
    \item \( P_{\text{no-aha}}(t) \) is the frequency proportion of token \( t \) among all no-aha tokens.
\end{itemize}

The sign and magnitude of the contrast probability quantify the bias of each token toward the style of one model: \( P_{\text{contrast}}(t) > 0 \) indicates that the token is more characteristic of the current category, while \( P_{\text{contrast}}(t) < 0 \) suggests that it is more representative of the other category. Ranging the contrast probabilities helps to identify the most distinctive tokens for each style.

\paragraph{Detailed statistical results of Aha token and no-aha token distribution.}
The statistical results, as shown in Figure \ref{fig:token_frequency}, reveal significant differences in the linguistic styles of aha tokens and no-aha tokens. Specifically, Aha tokens tend to adopt conversational and emotional expressions (e.g., 'wait', ``Alright,'' ``Hmm,'' ``So,'' ``Maybe''), making their linguistic style more similar to natural human expression when they are thinking. They exhibit strong interactivity and display anthropomorphic characteristics. This anthropomorphic style is preserved across models of different sizes and becomes more flexible as the size of the model increases. 

In models (1.5B), Aha tokens frequently use anthropomorphic expressions (e.g. ``Alright,'' ``I,'' ``Then''); in medium-scale models (7B), they gradually incorporate more logical and summarizing expressions (e.g., ``Finally,'' ``Thus''); and in large-scale models (14B), a better balance is achieved between anthropomorphism and logic. Additionally, more hypothetical expressions (e.g., ``Suppose,'' ``Perhaps'') are introduced, further enhancing the diversity and flexibility of the linguistic style.

In contrast, no-aha tokens emphasize logical and structured expressions (e.g. ``Let,'' ``However,'' ``To,'' ``This''), resulting in a linguistic style that appears more mechanical. Although it is clear and well organized, it lacks the natural and emotional qualities of human expression. As model size increases, the linguistic flexibility of no-aha tokens improves but remains focused on rigorous logical reasoning. For example, in small-scale models (1.5B), no-aha tokens frequently use operational and logical expressions (e.g., ``Let,'' ``To,'' ``Subtract''), leading to a relatively uniform style where reasoning relies heavily on explicit logical frameworks. In medium-scale (7B) and large-scale (14B) models, logical guiding expressions (e.g. ``Notice,'' ``Based'') are increasingly used, making the logical structure clearer. However, even in large-scale models (14B), their linguistic style continues to lack interactivity.

\paragraph{Anthropomorphic and reasoning token selection strategy.}

Based on the above statistical results and findings, we manually selected anthropomorphic tokens and reasoning tokens by observing the linguistic styles and contextual usage patterns of the tokens, ensuring that the results align with the distinctions between aha tokens and no-aha tokens. The results are shown in Table \ref{anthropomorphic_and_reasoning_tokens}, where these lists capture the different linguistic tendencies of the two models and are used for the calculation of our RASM metrics and analysis.

\begin{table}[h!]
\centering
\resizebox{\textwidth}{!}{
\begin{tabular}{>{\centering\arraybackslash}p{8cm}|>{\centering\arraybackslash}p{8cm}}
\toprule
\textbf{Anthropomorphic Tokens} & \textbf{Reasoning Tokens} \\ 
\midrule
"Aha", "I", "Hmm", "Oh", "Umm", "Well", "alright", "we", "Wait", "wait", "Alternatively", "Case", "Okay", "okay", "Perhaps", "Still", "we", "We", "Let", "Alright", "Good", "Consider", "Adjust", "Attempt", "Notice", "Try", "Test", "Using", "Oh", "Oops", "Break", "Looking", "Attempting", "Again", "Like", "Assume", "Maybe" & 
"This", "To", "Instead", "Therefore", "So", "As", "Because", "Since", "Thus", "However", "Calculate", "Notice", "Finally", "Multiply", "Divide", "Next", "Another", "First", "Second", "Third", "Final", "Simplifying", "Correct", "Incorrect", "Error", "combining", "Add", "Start", "Finally", "Conclusion", "Fourth", "Calcul" \\ 
\bottomrule
\end{tabular}
}
\caption{Selected anthropomorphic and reasoning tokens.}
\label{anthropomorphic_and_reasoning_tokens}
\end{table}







\section{Additional Results}
\subsection{Most probable yet uncertain sentences for each model}

We extracted and showcased the top 5 most uncertain sentences (with the lowest probability) for each model during the generation process, as shown in Tables \ref{tab:uncertainty_sentences_qwen_1.5b}, \ref{tab:uncertainty_sentences_qwen_7b}, \ref{tab:uncertainty_sentences_qwen_14b}, and \ref{tab:uncertainty_sentences_llama_8b}. For sentences that appear multiple times, we only consider the single instance with the lowest probability.

\label{app:uncertain sentences}
\begin{table}[h!]
    \centering
    \resizebox{\textwidth}{!}{
    \begin{tabular}{clclc}
        \toprule
        \textit{n} & \multicolumn{2}{c}{\textbf{Qwen2.5-Math-1.5B-Instruct}} & \multicolumn{2}{c}{\textbf{DeepSeek-R1-Distill-Qwen-1.5B}} \\
        \cmidrule(r){2-3} \cmidrule(l){4-5}
        & Sentences & Probability & Sentences & Probability \\
        \midrule
        \multicolumn{5}{c}{K\&K Task} \\
        \midrule
        3 & Identify and verify each... & 0.526006 & Hmm. & 0.054014 \\ 
          & Therefore, the solution is... & 0.632516 & That fits. & 0.152014 \\ 
          & This implies:\textbackslash n- & 0.665774 & That works. & 0.174358 \\ 
          & However,... & 0.676434 & Or maybe not. & 0.200675 \\ 
          & Based on the analysis:\textbackslash n... & 0.698401 & Maybe just list each one... & 0.268039 \\ 
        \midrule
        4 & Elizabeth said that ... & 0.514754 & Good. & 0.032638 \\ 
          & We will use logical reasoning... & 0.580071 & Hmm. & 0.120880 \\ 
          & Here's the reasoning:\textbackslash n\textbackslash n... & 0.596774 & That's settled. & 0.249905 \\ 
          & Final Conclusion: all... & 0.620608 & Is there another case? & 0.268122 \\ 
          & Harper's statement\textbackslash n & 0.622434 & So, that seems to... & 0.287400 \\ 
        \midrule
        6 & Let's consider these cases... & 0.416611 & Hmm. & 0.058834 \\ 
          & Then:\textbackslash n- & 0.511978 & Let's check back. & 0.196249 \\ 
          & This means Benjamin cannot be... & 0.630602 & Maybe they're different people... & 0.210393 \\ 
          & Let's check the implications... & 0.633654 & But let's see:\textbackslash n\textbackslash n... & 0.268735 \\ 
          & Amelia's identity (k... & 0.667910 & Maybe I need to represent... & 0.279453 \\ 
        \midrule
        8 & Governing the gap... & 0.426626 & Hmm. & 0.045391 \\ 
          & Ben only.\textbackslash n- & 0.484353 & So, the paradox arises... & 0.291671 \\ 
          & We notice that the statement... & 0.510301 & Got that down. & 0.295538 \\ 
          & This is a contradiction. & 0.561809 & So, maybe the remaining... & 0.343974 \\ 
          & O budget: a knight... & 0.582986 & Then, is that possible... & 0.348499 \\ 
        \midrule
        10 & Therefore:\textbackslash n- & 0.373919 & Hmm. & 0.175674 \\ 
          & guiding the statement of guided... & 0.503938 & Or is Logan another? & 0.254154 \\ 
          & Joseph's statement is consistent... & 0.549551 & Wait, this is a... & 0.304910 \\ 
          & Let's summarize:\textbackslash n\textbackslash n- & 0.550420 & Then, since ** & 0.309152 \\ 
          & Let \( K \)... & 0.577642 & Or perhaps there was a... & 0.318907 \\ 
        \midrule
        \multicolumn{5}{c}{Count Down Task} \\
        \midrule
       3 & One approach is to use... & 0.662153 & I'm really stuck. & 0.219872 \\ 
          & We are there! & 0.670710 & So, negative numbers might... & 0.290523 \\ 
          & First, we notice that... & 0.688039 & Let's think about combining... & 0.311512 \\ 
          & Adjust the approach by trying... & 0.697551 & Maybe using the digits in... & 0.334947 \\ 
          & First, consider the operation... & 0.706210 & Maybe subtract something? & 0.342023 \\ 
        \midrule
        4 & Consider using addition and multiplication... & 0.518555 & Hmm. & 0.079202 \\ 
          & Start by identifying which operations... & 0.614555 & Oh! & 0.150123 \\ 
          & Consider the number 7... & 0.667058 & Oops! & 0.174373 \\ 
          & However, this doesn't... & 0.686095 & Nope. & 0.275300 \\ 
          & We can subtract 2... & 0.703876 & This seems like a step... & 0.364554 \\ 
        \midrule
        6 & Start by identifying possible combinations... & 0.624272 & Hmm. & 0.017172 \\ 
          & Let's see if we... & 0.664639 & Maybe ... & 0.041119 \\ 
          & Let's instead try subtract... & 0.669138 & But what? & 0.084418 \\ 
          & First, we try to... & 0.678575 & Perfect! & 0.165579 \\ 
          & We can use the other... & 0.708372 & Oh! & 0.182604 \\ 
        \midrule
        8 & Identify the numbers and possible... & 0.484072 & Hmm & 0.036037 \\ 
          & Identify potential combinations and operations... & 0.622140 & Maybe subtract somewhere? & 0.119820 \\ 
          & This... & 0.642424 & Not helpful & 0.147793 \\ 
          & Consider the number 9... & 0.642484 & That's good. & 0.228477 \\ 
          & Identify the largest numbers and... & 0.653467 & Let's think about that... & 0.258753 \\ 
        \midrule
        10 & Identify potential combinations of numbers... & 0.534293 & Hmm. & 0.022847 \\ 
          & Identify the largest number and... & 0.600052 & Perfect. & 0.041693 \\ 
          & We can use the numbers... & 0.604406 & Oh! & 0.115664 \\ 
          & We need to experiment with... & 0.610839 & Let's try. & 0.146562 \\ 
          & Another approach is to look... & 0.637062 & Maybe that could work? & 0.195875 \\ 
        \bottomrule
    \end{tabular}
    }
    \caption{Top 5 probable yet uncertain sentences (the first 5 tokens) generated by Qwen2.5-Math-1.5B-Instruct and DeepSeek-R1-Distill-Qwen-7B in different difficulty-level. For sentences that appear multiple times, we only consider the single instance with the lowest probability.}
    \label{tab:uncertainty_sentences_qwen_1.5b}
\end{table}

\begin{table}[h!]
    \centering
    \resizebox{\textwidth}{!}{
    \begin{tabular}{clclc}
        \toprule
        \textit{n} & \multicolumn{2}{c}{\textbf{Qwen2.5-Math-7B-Instruct}} & \multicolumn{2}{c}{\textbf{DeepSeek-R1-Distill-Qwen-7B}} \\
        \cmidrule(r){2-3} \cmidrule(l){4-5}
        & Sentences & Probability & Sentences & Probability \\
        \midrule
        \multicolumn{5}{c}{K\&K Task} \\
        \midrule
        3 & If given a unique answer... & 0.021290 & Hmm. & 0.075861 \\ 
          & Charlotte is not a kn... & 0.022217 & Wait & 0.132202 \\ 
          & Let's re-evaluate... & 0.022454 & Let me verify again:\textbackslash n\textbackslash n... & 0.192162 \\ 
          & Revisiting the problem... & 0.023387 & Wait, no. & 0.320067 \\ 
          & Therefore, Victoria cannot be... & 0.024983 & Hmm, not sure. & 0.343187 \\ 
        \midrule
        4 & This contradictionocalypse already ... & 0.020467 & Hmm. & 0.025640 \\ 
          & Based on given logical logical... & 0.021195 & Maybe both. & 0.226319 \\ 
          & We need to find the... & 0.021396 & Let me think. & 0.260847 \\ 
          & Therefore, let's re... & 0.021410 & Wait a second! & 0.271614 \\ 
          & Therefore, there is a... & 0.021418 & But we have the implication... & 0.283957 \\ 
        \midrule
        6 & This contradiction suggests reevaluation... & 0.021000 & Interesting. & 0.145864 \\ 
          & Therefore, we reector... & 0.021206 & Hmm. & 0.163861 \\ 
          & Since both are always true... & 0.021938 & Let me reconsider. & 0.314669 \\ 
          & Therefore, Jack must be... & 0.022510 & No, no. & 0.364315 \\ 
          & Thus, based on the... & 0.023170 & I just thought of another... & 0.382149 \\ 
        \midrule
        8 & Let's againinateuellen... & 0.021018 & Hmm. & 0.222903 \\ 
          & Daniel statement... & 0.021364 & Interesting. & 0.238739 \\ 
          & Therefore, Noah must be... & 0.023305 & So, is that consistent... & 0.326876 \\ 
          & However, since we have... & 0.025103 & That means:\textbackslash n\textbackslash n- & 0.339934 \\ 
          & However, based on logical... & 0.029722 & Owen is knight? & 0.340216 \\ 
        \midrule
        10 & Charlotte\textbackslash n\textbackslash nHowever, re... & 0.021429 & Hmm. & 0.090210 \\ 
          & Elizabeth is a knight Lucas... & 0.021480 & We know that ... & 0.163797 \\ 
          & This contradiction indicates a logical... & 0.021510 & Let's re-express... & 0.220035 \\ 
          & Olivia: knight ( contradict... & 0.021734 & However, we know from... & 0.255357 \\ 
          & If Aria is indeed... & 0.021865 & Let me also consider James... & 0.332764 \\ 
        \midrule
        \multicolumn{5}{c}{Count Down Task} \\
        \midrule
       3 & scribbola searchlegate... & 0.020352 & I'll consider using division... & 0.475217 \\ 
          & then it subtract we... & 0.020729 & I need a way to... & 0.487473 \\ 
          & Finally, we may... & 0.020836 & This is getting tricky... & 0.505426 \\ 
          & we need to... & 0.020858 & This works perfectly. & 0.526299 \\ 
          & Finally we need... & 0.020544 & To refine this,... & 0.546421 \\ 
        \midrule
        4 & Correct calculation:\textbackslash n\textbackslash n... & 0.020353 & Wait a minute. & 0.422414 \\ 
          & Letisors(coefficients)value... & 0.020503 & Nope. & 0.425691 \\ 
          & 57+23... & 0.020666 & Let's start by exploring... & 0.462525 \\ 
          & so we can work it... & 0.020936 & We'll explore the following... & 0.488311 \\ 
          & Next, find another approach... & 0.021074 & Wait, this gives ... & 0.530180 \\ 
        \midrule
        6 & Finally we need... & 0.020544 & Hmm. & 0.018334 \\ 
          & this approach can be... & 0.020800 & No. & 0.142961 \\ 
          & Let's use:\textbackslash n\textbackslash n & 0.020831 & Oh! & 0.191606 \\ 
          & If we duty seed box... & 0.020854 & Consider using ... & 0.194675 \\ 
          & consider the number 2... & 0.020924 & Let me try that. & 0.215487 \\ 
        \midrule
        8 & We may finish with ... & 0.020462 & Nope. & 0.038214 \\ 
          & Subtract:\textbackslash n\textbackslash n & 0.020575 & Hmm. & 0.122892 \\ 
          & After furtheranging and checking... & 0.020636 & Hmm & 0.155568 \\
          & 8 + ... & 0.020742 & But how? & 0.191274 \\ 
          & let's add revolutions.\textbackslash n\textbackslash n... & 0.020858 & No. & 0.193985 \\ 
        \midrule
        10 & Re consider basic ... & 0.020173 & Perfect. & 0.029956 \\ 
          & 255 adult... & 0.020467 & But that still seems unclear... & 0.060331 \\ 
          & Let us try witnessed ... & 0.020590 & Hmm. & 0.078308 \\ 
          & Let's use:\textbackslash n\textbackslash n & 0.020845 & Not sure. & 0.114357 \\ 
          & The previous unilateral way I... & 0.020947 & Maybe combining some operations. & 0.167437 \\ 
        \bottomrule
    \end{tabular}
    }
    \caption{Top 5 probable yet uncertain sentences (the first 5 tokens) generated by Qwen2.5-Math-7B-Instruct and DeepSeek-R1-Distill-Qwen-7B in different difficulty-level. For sentences that appear multiple times, we only consider the single instance with the lowest probability.}
    \label{tab:uncertainty_sentences_qwen_7b}
\end{table}

\begin{table}[h!]
    \centering
    \resizebox{\textwidth}{!}{
    \begin{tabular}{clclc}
        \toprule
        \textit{n} & \multicolumn{2}{c}{\textbf{Qwen2.5-14B-Instruct}} & \multicolumn{2}{c}{\textbf{DeepSeek-R1-Distill-Qwen-14B}} \\
        \cmidrule(r){2-3} \cmidrule(l){4-5}
        & Sentences & Probability & Sentences & Probability \\
        \midrule
        \multicolumn{5}{c}{K\&K Task} \\
        \midrule
        3 & Because the statement is self... & 0.560364 & Hmm. & 0.236663 \\ 
          & We have:\textbackslash n- & 0.581820 & Let's be careful here... & 0.296178 \\ 
          & Therefore, both cannot be... & 0.604704 & Therefore, contradiction. & 0.328875 \\ 
          & This contradiction confirms our earlier... & 0.673183 & Is that acceptable? & 0.374225 \\ 
          & Charlotte is a knight.\textbackslash n\textbackslash n... & 0.674642 & verify again. & 0.391501 \\ 
        \midrule
        4 & Therefore:\textbackslash n- & 0.415222 & Hmm. & 0.086689 \\ 
          & Samuel: Knight\textbackslash n & 0.604418 & Let me think. & 0.328805 \\ 
          & Given all statements and logical... & 0.607570 & Or can it be either... & 0.355387 \\ 
          & But since we need consistency... & 0.614379 & Let me consider the possibilities... & 0.393884 \\ 
          & However, since Jacob claims... & 0.618115 & Maybe like this:\textbackslash n\textbackslash n- & 0.395614 \\ 
        \midrule
        6 & This would mean Isabella... & 0.496426 & Hmm. & 0.064495 \\ 
          & This contradiction means our assumption... & 0.517584 & Let's think carefully. & 0.247777 \\ 
          & Based on all consistent statements... & 0.539659 & Let's check that. & 0.296434 \\ 
          & But Owen is a knight... & 0.555271 & Let me clarify. & 0.305299 \\ 
          & Charlotte as Knight: True... & 0.580151 & All statements hold. & 0.319862 \\ 
        \midrule
        8 & James: Not specified in... & 0.378060 & Hmm. & 0.073389 \\ 
          & But for completeness:\textbackslash n- & 0.414346 & Let's see. & 0.220944 \\ 
          & This means Samuel is a... & 0.545051 & That's odd. & 0.309094 \\ 
          & Aurora: Knight\textbackslash n & 0.564745 & Which is already true. & 0.338379 \\ 
          & James: Can be either... & 0.579905 & Let's see how they... & 0.404691 \\ 
        \midrule
        10 & This doesn't give us... & 0.518703 & Hmm. & 0.154247 \\ 
          & Olivia's status is still... & 0.596697 & Is that consistent? & 0.309588 \\ 
          & Hence, let's proceed... & 0.616119 & So that works. & 0.320555 \\ 
          & Elizabeth: Unknown for now... & 0.624314 & Let me rephrase. & 0.334192 \\ 
          & Joseph must be a kn... & 0.639972 & Let's clarify. & 0.338246 \\ 
        \midrule
        \multicolumn{5}{c}{Count Down Task} \\
        \midrule
       3 & We need to re-e... & 0.544595 & Hmm. & 0.070761 \\ 
          & We need to get creative... & 0.553488 & Nope. & 0.101929 \\ 
          & We need to think about... & 0.566548 & Not helpful. & 0.149741 \\ 
          & We need to rethink our... & 0.569159 & Maybe addition? & 0.286247 \\ 
          & After trying various combinations,... & 0.577618 & Nope, still way too... & 0.359081 \\ 
        \midrule
        4 & After several attempts,... & 0.490595 & Still & 0.080027 \\ 
          & Let’s verify again carefully... & 0.509019 & Hmm. & 0.090551 \\ 
          & Rechecking calculations and ensuring... & 0.556210 & How? & 0.131233 \\ 
          & It seems... & 0.631089 & Let me try. & 0.198806 \\ 
          & Here’s a step-by... & 0.644402 & All right. & 0.213726 \\ 
        \midrule
        6 & Let's recheck our... & 0.382034 & Hmm. & 0.100985 \\ 
          & We need a better combination... & 0.406203 & Perfect! & 0.121031 \\ 
          & But this approach should work... & 0.484926 & Not quite. & 0.247189 \\ 
          & This confirms the... & 0.536346 & But maybe subtract instead. & 0.254818 \\ 
          & We need to re-e... & 0.587170 & Let me see. & 0.256791 \\ 
        \midrule
        8 & It might require rechecking... & 0.358824 & Exactly! & 0.046170 \\ 
          & Use multiplication or addition involving... & 0.417593 & Yes! & 0.052473 \\ 
          & After several attempts,... & 0.546278 & Hmm. & 0.063112 \\ 
          & We need to find a... & 0.571972 & Perfect! & 0.075026 \\ 
          & After verifying... & 0.579883 & Ah! & 0.151318 \\ 
        \midrule
        10 & The best approach is to... & 0.496744 & Hmm. & 0.031313 \\ 
          & Given the constraints, we... & 0.591498 & Still over. & 0.101248 \\ 
          & But instead of... & 0.621007 & Not enough. & 0.122928 \\ 
          & This attempt didn... & 0.634036 & Maybe not. & 0.128383 \\ 
          & After several attempts, it... & 0.640827 & Let me think. & 0.224669 \\ 
        \bottomrule
    \end{tabular}
    }
    \caption{Top 5 probable yet uncertain sentences (the first 5 tokens) generated by Qwen2.5-14B-Instruct and DeepSeek-R1-Distill-Qwen-14B in different difficulty-level. For sentences that appear multiple times, we only consider the single instance with the lowest probability.}
    \label{tab:uncertainty_sentences_qwen_14b}
\end{table}

\begin{table}[h!]
    \centering
    \resizebox{\textwidth}{!}{
    \begin{tabular}{clclc}
        \toprule
        \textit{n} & \multicolumn{2}{c}{\textbf{Llama-3.1-8B-Instruct}} & \multicolumn{2}{c}{\textbf{DeepSeek-R1-Distill-Llama-8B }} \\
        \cmidrule(r){2-3} \cmidrule(l){4-5}
        & Sentences & Probability & Sentences & Probability \\
        \midrule
        \multicolumn{5}{c}{K\&K Task} \\
        \midrule
        3 & From this, we... & 0.420664 & Hmm. & 0.134315 \\ 
          & This means this statement is... & 0.434759 & So this seems impossible. & 0.185909 \\ 
          & This would create another paradox... & 0.484552 & That seems redundant. & 0.253995 \\ 
          & Now, let's... & 0.510904 & So, Riley is K... & 0.271537 \\ 
          & To solve the puzzle properly... & 0.528183 & Maybe more precise:\textbackslash n\textbackslash n- & 0.299061 \\ 
        \midrule
        4 & However, the initial... & 0.375455 & Hmm. & 0.028437 \\ 
          & This implies that our assumption... & 0.401252 & Interesting. & 0.111463 \\ 
          & But, let's... & 0.401558 & Huh. & 0.241335 \\ 
          & Let's continue with the... & 0.522880 & Not sure. & 0.244895 \\ 
          & However, we can resolve... & 0.537905 & But wait, no—... & 0.268583 \\ 
        \midrule
        6 & Now let's check... & 0.479586 & Hmm. & 0.059771 \\ 
          & So, this statement is... & 0.488198 & Let me think. & 0.140541 \\ 
          & However, considering all... & 0.495336 & It's possible. & 0.240117 \\ 
          & Now we know all the... & 0.526281 & That helps us. & 0.248701 \\ 
          & Chloe: Knight & 0.540745 & Is that the only possibility... & 0.297448 \\ 
        \midrule
        8 & This would make Mia a... & 0.444089 & Hmm. & 0.098603 \\ 
          & However, this contradiction arises... & 0.466709 & Interesting. & 0.260285 \\ 
          & The last two statements are... & 0.467839 & But wait, no,... & 0.312512 \\ 
          & Scarlett is not mentioned in... & 0.498923 & So let me summarize:\textbackslash n\textbackslash n... & 0.324689 \\ 
          & This leaves us with ... & 0.524408 & That works. & 0.331900 \\ 
        \midrule
        10 & This would make both statements... & 0.331598 & Hmm. & 0.038465 \\ 
          & Now we can conclude... & 0.464352 & That might be important. & 0.196700 \\ 
          & Knights: Only 7... & 0.465627 & Or maybe not. & 0.210730 \\ 
          & This creates a paradox. & 0.475077 & Let's see. & 0.249268 \\ 
          & This statement is false. & 0.485319 & Wait, contradiction. & 0.294464 \\ 
        \midrule
        \multicolumn{5}{c}{Count Down Task} \\
        \midrule
       3 & Now, we need to... & 0.352150 & Maybe. & 0.039438 \\ 
          & However, we need... & 0.413969 & Hmm & 0.069371 \\ 
          & 32 / 0... & 0.467406 & Hmm. & 0.168900 \\ 
          & So, let's try... & 0.470556 & Maybe division. & 0.230316 \\ 
          & Let's use the... & 0.490372 & How? & 0.257203 \\ 
        \midrule
        4 & However, that would require... & 0.343084 & Hmm. & 0.016199 \\ 
          & 57, close but not... & 0.376949 & No. & 0.032682 \\ 
          & Start with 88\textbackslash n & 0.425079 & Oh! & 0.047348 \\ 
          & Start by trying to combine... & 0.495647 & That works! & 0.156867 \\ 
          & Using 6. & 0.495879 & Maybe subtract instead. & 0.160207 \\ 
        \midrule
        6 & 693 + 5... & 0.417244 & No. & 0.025702 \\ 
          & However, that is not... & 0.436204 & Yes! & 0.026157 \\ 
          & 26=67 and ... & 0.493914 & How? & 0.054781 \\ 
          & But we need to get... & 0.507572 & Alternatively & 0.056994 \\ 
          & 25 is not possible,... & 0.528331 & Oh! & 0.177169 \\ 
        \midrule
        8 & 42\textbackslash n & 0.350759 & How? & 0.027835 \\ 
          & 121\textbackslash n & 0.400639 & Maybe. & 0.037071 \\ 
          & 605\textbackslash n & 0.453578 & Hmm. & 0.050594 \\ 
          & Now we need to... & 0.491798 & Same result. & 0.087146 \\ 
          & However,... & 0.497138 & 25? & 0.113435 \\ 
        \midrule
        10 & 676\textbackslash n & 0.117458 & Maybe. & 0.025831 \\ 
          & 8889. & 0.273806 & No. & 0.043118 \\ 
          & 75, so that gives... & 0.324264 & Hmm. & 0.051259 \\ 
          & However, we can also... & 0.336813 & Ah! & 0.133850 \\ 
          & 5 either. & 0.382819 & Close. & 0.155760 \\ 
        \bottomrule
    \end{tabular}
    }
    \caption{Top 5 probable yet uncertain sentences (the first 5 tokens) generated by Llama-3.1-8B-Instruct and DeepSeek-R1-Distill-Llama-8B in different difficulty-level. For sentences that appear multiple times, we only consider the single instance with the lowest probability.}
    \label{tab:uncertainty_sentences_llama_8b}
\end{table}





\subsection{Additional results for Reasoning-Anthropomorphic Separation}
\label{app:Additional results for Reasoning-Anthropomorphic Separation}
\begin{figure*}[t]
\centering
\includegraphics[width=\textwidth]{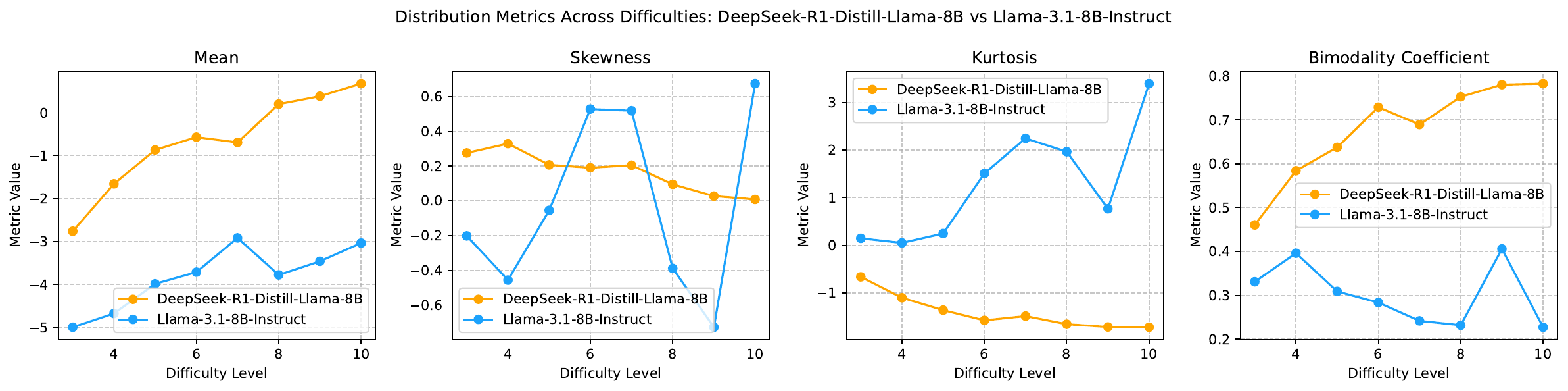}
\includegraphics[width=\textwidth]{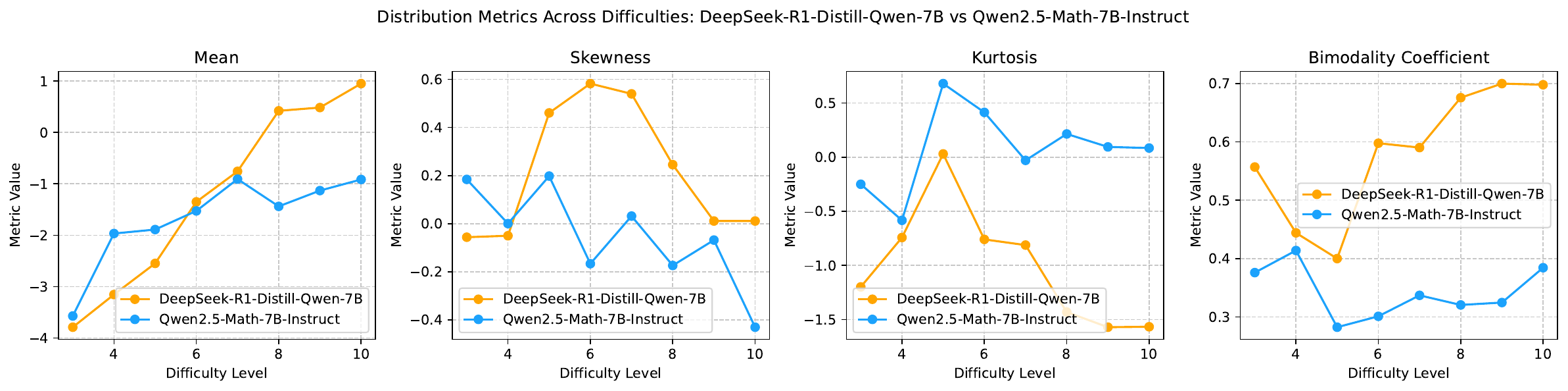}
\includegraphics[width=\textwidth]{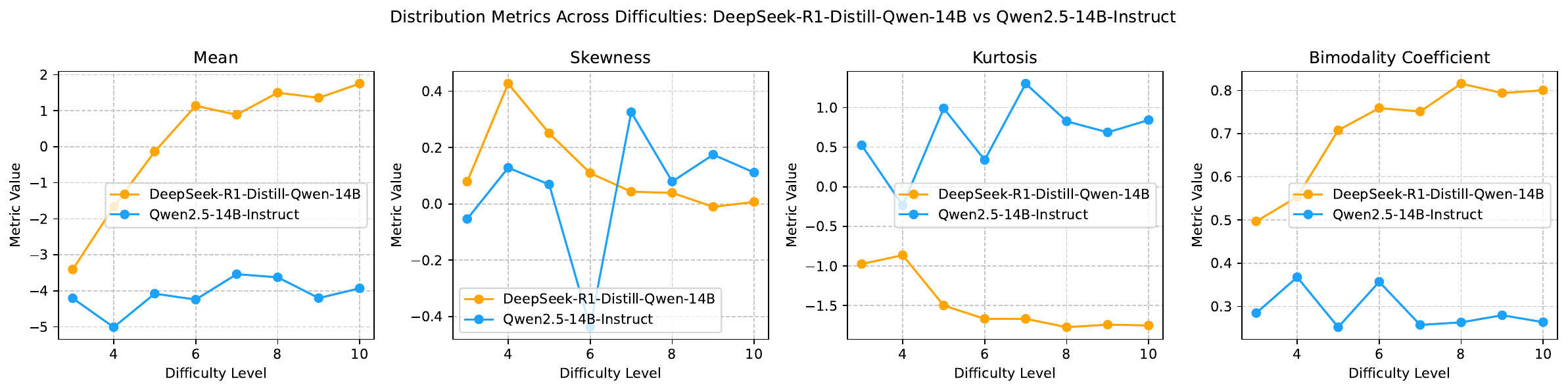}
\includegraphics[width=\textwidth]{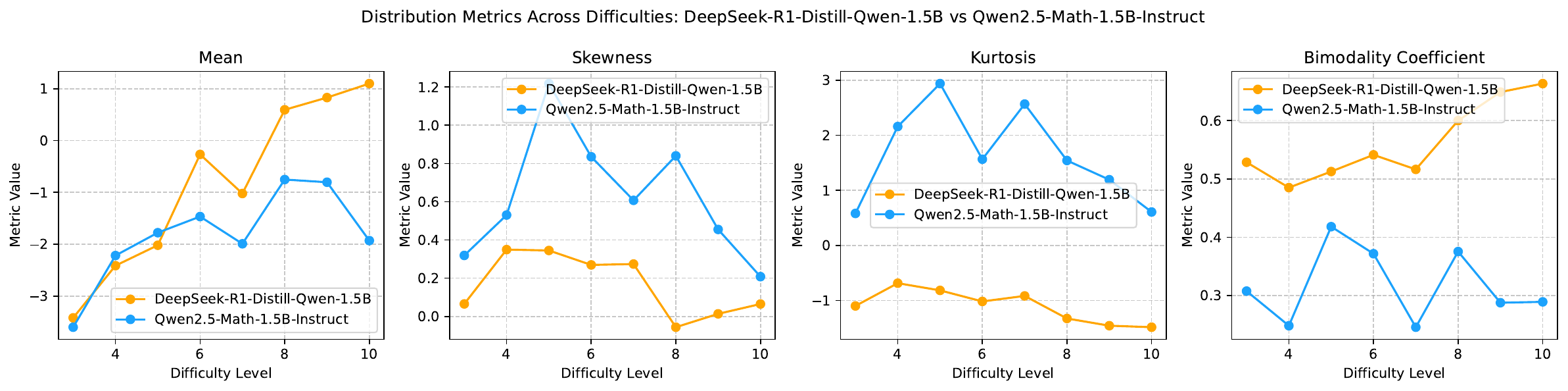}
\caption{Distribution Metrics Across Difficulties} 
\label{fig: Distribution Metrics}
\end{figure*}

In Figure~\ref{fig: Distribution Metrics}, we present the additional metric for Reasoning-Anthropomorphic Separation (RASM) discussed in Section~\ref{subsec:rasm}.

\end{document}